%% file: main.tex
\documentclass[10pt,twocolumn,letterpaper]{article}

\usepackage[pagenumbers]{cvpr} % To force page numbers, e.g. for an arXiv version

\input{preambles/macro}

\input{preambles/math_commands.tex}

\input{preambles/def}

\usepackage{tikz}
\usepackage{wrapfig}

\definecolor{cvprblue}{rgb}{0.21,0.49,0.74}
\usepackage[pagebackref,breaklinks,colorlinks,allcolors=cvprblue]{hyperref}
\usepackage{bbm}

\def\paperID{12970} % *** Enter the Paper ID here
\def\confName{CVPR}
\def\confYear{2025}

\title{
Dr. Splat: Directly Referring 3D Gaussian Splatting \\ via Direct Language Embedding Registration
}

\input{preambles/authors}

\begin{document}
\maketitle

\input{section/0_abstract}
\input{section/1_intro}
\input{section/2_related_works}
\input{figures/2-overview/2-overview}
\input{section/3_preliminary}
\input{figures/4-pq/4-pq}
\input{section/4_0_method}
\input{section/4_1}

\input{tables/Object_selection}
\input{figures/5-obj_sel/5-obj_sel}
\input{section/4_2}
\input{tables/3d_loc_and_seg}

\input{figures/7-weight_verification/7_weight_verification}
\input{figures/6-loc_and_seg/6-loc}
\input{section/4_3}
\input{section/5_0_experiment}

\input{section/5_1}

\input{section/5_2}
\input{section/5_3}
\input{section/5_4}

\input{figures/8-ablation/two_graphs}
\input{section/6_conclusion}

{
    \small
    \bibliographystyle{ieeenat_fullname}
    \bibliography{main}
}

\clearpage
\appendix
\maketitlesupplementary

{
\hypersetup{linkcolor=black}
\noindent\hyperref[sec:implementation]{\textbf{A \ \ \ Implementation Details}}\\
\hyperref[sec:exp_setup]{\textbf{B \ \ \ Experiment Setup}} \\ 
\hyperref[metric]{\textbf{C \ \ \ Evaluation Protocols}} \\ 
\hyperref[pq]{\textbf{D \ \ \ Search-time Experiments }} \\ 
\hyperref[additional results]{\textbf{E \ \ \ Additional Results}} \\ 
\hyperref[3D tasks]{\text{\qquad E.1 \ \ \ Additional results on presented 3D tasks}} \\ 
\hyperref[scannet200]{\text{\qquad E.2 \ \ \ Experiments on the ScanNet200 dataset}} \\ 
\hyperref[city scale]{\text{\qquad E.3 \ \ \ Experiments on the city-scale dataset}} \\ 
\hyperref[discussion]{\textbf{F \ \ \ Broader Applications and Limitations }} \\ 
}
\noindent\rule{\linewidth}{0.2pt}

\setcounter{figure}{0}
\renewcommand{\thefigure}{S\arabic{figure}}
\renewcommand{\thetable}{S\arabic{table}}
\appendix
\input{supp/0_head}
\input{supp_figures/S1_pq}
\input{supp/1.implementation}

\input{supp_figures/S2_metric_qualitative}
\input{supp/2.experiment_setup}
\input{supp/3.1.prev_eval_protocol.tex}

\input{supp/3.2.our_eval_protocol}
\input{supp_tables/metric_comparison}

\input{supp_figures/S3_metric_correlation}
\input{supp/4.pq}
\input{supp/5_additional_results}
\input{supp/6_limitation}

\end{document}

%% file: preambles/macro.tex
\newcommand{\nickname}{Dr.~Splat\xspace}

\makeatletter
\DeclareRobustCommand\onedot{\futurelet\@let@token\@onedot}
\def\@onedot{\ifx\@let@token.\else.\null\fi\xspace}
\def\eg{\emph{e.g}\onedot} 
\def\ie{\emph{i.e}\onedot}

\newcommand{\noindentbold}[1]{\noindent \textbf{#1.}\xspace}

\newcommand{\featmap}{\mathbf{f}^\text{map}}

%% file: preambles/math_commands.tex
\usepackage{amsmath,amsfonts,bm}

\def\eqref#1{equation~\ref{#1}}
\def\1{\bm{1}}

\DeclareMathAlphabet{\mathsfit}{\encodingdefault}{\sfdefault}{m}{sl}
\SetMathAlphabet{\mathsfit}{bold}{\encodingdefault}{\sfdefault}{bx}{n}

%% file: preambles/def.tex
\usepackage{enumitem}
\usepackage{multirow}
\usepackage{colortbl}
\usepackage{microtype}
\usepackage{comment}
\usepackage[hang,flushmargin]{footmisc}

\usepackage{amsfonts}

\setlist[itemize]{align=parleft,left=0pt,topsep=1mm,itemsep=0mm,parsep=1mm}

\definecolor{azure(colorwheel)}{rgb}{0.0, 0.5, 1.0}
\definecolor{R5}{rgb}{0.0, 0.7, 0.1}
\definecolor{yw}{rgb}{0.01176, 0.5490, 0.5490}
\definecolor{R123}{rgb}{0.36, 0.54, 0.66}
\definecolor{R1234}{rgb}{0.7, 0.75, 0.71}
\definecolor{applegreen}{rgb}{0.55, 0.71, 0.0}
\definecolor{R132}{rgb}{0.0, 0.0, 1.0}
\definecolor{postechred}{rgb}{0.784, 0.003, 0.313}
\definecolor{gu}{rgb}{0.5460, 0.1755, 0.2766}
\definecolor{el}{rgb}{0.9764, 0.447, 0.447}
\definecolor{hyos}{rgb}{0.662, 0.482, 0.960}
\definecolor{ballblue}{rgb}{0.13, 0.67, 0.8}
\definecolor{cornellred}{rgb}{0.7, 0.11, 0.11}
\definecolor{darkcyan}{rgb}{0.0, 0.55, 0.55}
\definecolor{CuGray}{gray}{0.9}
\definecolor{airforceblue}{rgb}{0.36, 0.54, 0.66}
\definecolor{rev}{rgb}{0.784, 0.003, 0.313}
\definecolor{pink}{cmyk}{0, 0.7808, 0.4429, 0.1412}
\definecolor{amethyst}{rgb}{0.6, 0.4, 0.8}
\definecolor{black}{rgb}{0.0, 0.0, 0.0}
\definecolor{tb3_yellow}{rgb}{0.996, 1.0, 0.6}
\definecolor{R123}{rgb}{0.980, 0.8, 0.604}
\definecolor{R512}{rgb}{0.972, 0.6, 0.6}
\definecolor{dimgray}{rgb}{0.41, 0.41, 0.41}
\definecolor{R3}{rgb}{0.8, 0.25, 0.33}
\definecolor{bleudefrance}{rgb}{0.19, 0.55, 0.91}
\definecolor{R6}{rgb}{0.265, 0.445, 0.765}
\definecolor{blue(ryb)}{rgb}{0.01, 0.28, 1.0}
\definecolor{R4}{rgb}{1.0, 0.49, 0.0}
\definecolor{Gray}{gray}{0.88}
\definecolor{green(ncs)}{rgb}{0.0, 0.62, 0.42}
\definecolor{brightpink}{rgb}{1.0, 0.0, 0.5}
\definecolor{alizarin}{rgb}{0.82, 0.1, 0.26}
\definecolor{clova}{rgb}{0.24, 0.63, 0.33}
\definecolor{orange-red}{rgb}{1.0, 0.27, 0.0}
\definecolor{nicegreen}{rgb}{0.0, 0.7, 0.1}

\definecolor{kellygreen}{rgb}{0.3, 0.73, 0.09}

\newcolumntype{g}{>{\columncolor{CuGray}}c}
\newcolumntype{z}{>{\columncolor{CuGray}}l}

\renewcommand{\paragraph}[1]{\vspace{1mm}\noindent\textbf{#1.}\,}

\usepackage{xspace}

\makeatletter
\def\@fnsymbol#1{\ensuremath{\ifcase#1\or *\or \dagger\or \ddagger\or
   \mathsection\or \mathparagraph\or \|\or **\or \dagger\dagger
   \or \ddagger\ddagger \else\@ctrerr\fi}}
\makeatother

\def\onedot{.\@\xspace}
\def\eg{\emph{e.g}\onedot} 
\def\ie{\emph{i.e}\onedot}

\newcommand{\Sref}[1]{Sec.~\ref{#1}}
\newcommand{\Eref}[1]{Eq.~(\ref{#1})}
\newcommand{\Fref}[1]{Fig.~\ref{#1}}
\newcommand{\Tref}[1]{Table~\ref{#1}}

\newcommand{\btheta}{\mbox{\boldmath $\theta$}}

\newcommand{\bmu}{\mbox{\boldmath $\mu$}}

\newcommand{\be}{\begin{eqnarray}}
\newcommand{\ee}{\end{eqnarray}}
\newcommand{\bee}{\begin{eqnarray*}}
\newcommand{\eee}{\end{eqnarray*}}

\newcommand{\matrixb}{\left[ \begin{array}}
\newcommand{\matrixe}{\end{array} \right]}

\usepackage{amssymb}
\usepackage{pifont}
\usepackage{algorithm}
\usepackage{algpseudocode}

%% file: preambles/authors.tex
\author{
Kim Jun-Seong$^{1}$ \qquad
GeonU Kim$^{1}$ \qquad
Kim Yu-Ji$^{1}$ \qquad 
Yu-Chiang Frank Wang$^{2}$ \\
Jaesung Choe$^{2}$\thanks{Corresponding Authors} \qquad 
Tae-Hyun Oh$^{3\ast}$ \\
\\
{$^{1}$POSTECH} \quad
{$^{2}$NVIDIA}  \quad
{$^{3}$KAIST}  \quad
\vspace{-4mm}
}

%% file: section/0_abstract.tex
\begin{abstract}

We introduce \nickname, a novel approach for open-vocabulary 3D scene understanding leveraging 3D Gaussian Splatting. Unlike existing language-embedded 3DGS methods, which rely on a rendering process, our method directly associates language-aligned CLIP embeddings with 3D Gaussians for holistic 3D scene understanding. The key of our method is a language feature registration technique where CLIP embeddings are assigned to the dominant Gaussians intersected by each pixel-ray. Moreover, we integrate Product Quantization (PQ) trained on general large-scale image data to compactly represent embeddings without per-scene optimization. Experiments demonstrate that our approach significantly outperforms existing approaches in 3D perception benchmarks, such as open-vocabulary 3D semantic segmentation, 3D object localization, and 3D object selection tasks. 
For video results, please visit : 
\url{https://drsplat.github.io/}

\end{abstract}

%% file: section/1_intro.tex
\section{Introduction}
\label{sec:introduction}

Open-vocabulary 3D scene understanding represents a significant challenge in the field of computer vision, with applications spanning autonomous navigation, robotics, and augmented reality. 
This approach aims to enable the interpretation and referencing of 3D spatial information through natural language, allowing for applicability beyond a restricted set of predefined categories~\cite{point_transformer,point_mixer,point_next,mink,point_net,point_transformer_v3,super_point_trandsformer}. 
Previously, open-vocabulary 3D scene understanding has been explored using point-cloud-based methods~\cite{open_scene,regionplc,ov3d,openmask3d,openins3d,langsplat,lerf}. 
Recently, the 3D Gaussian Splatting (3DGS) ~\cite{3dgs} has introduced a continuous representation integrated on explicit 3D Gaussians, which differs from traditional point-cloud approaches, enabling rapid progress in practical applications~\cite{gaussian_grasper}. 
Current research has begun to explore methods for associating language-based features with 3D Gaussian splats to enhance scene understanding capabilities.

\input{figures/1-teaser/teaser}

Several recent approaches~\cite{langsplat, legaussian, fmgs} introduce 3D Gaussian representation~\cite{3dgs} into the open-vocabulary scene understanding. 
This unique representation uses 3D Gaussians to achieve high-quality scene rendering, offering a more structured representation that addresses some limitations of point clouds. 
Building on this, these methods employ 2D~vision-language models to transfer language knowledge to 3D Gaussians ``via rendered feature maps''.

Despite its promise, such rendering-based distillation methods~\cite{langsplat, legaussian} share two limitations. First, we found that there is a discrepancy between optimized embeddings in 3D Gaussians and 2D language-aligned embeddings. This gap arises mainly from an intermediate rendering step that may distort CLIP embeddings during training. Then, the reliance on rendering impedes holistic 3D scene understanding, additional task-processing such as 3D semantic segmentation and 3D object localization, and making full spatial coverage calculations less efficient than direct 3D Gaussian methods~\cite{open_gaussian} including ours as illustrated in~\Fref{fig:teaser}.

To address this issue, this work proposes \nickname. Our method bypasses the rendering stage, enabling direct interaction with 3D Gaussians for registering and referring the well-preserved language-aligned CLIP embeddings in the 3D space. This makes our \nickname clearly distinguishable from prior works, facilitating a seamless integration of representative embeddings from 2D vision language models into the 3D spatial structure without compromising exhaustive rendering process that has been exploited~\cite{langsplat, legaussian, zhou2024feature, fastlgs, fmgs, gaussian_grasper}. Moreover, we propose to use a Product Quantization (PQ) feature encoding method to represent embeddings compactly and efficiently without any per-scene optimization. Rather than storing full-length feature vectors or per-scene specifically compressed embeddings~\cite{langsplat, legaussian, zhou2024feature, fastlgs, fmgs, gaussian_grasper}, each Gaussian in our \nickname stores an index from a pre-trained PQ, significantly reducing memory usage up to 6.25$\%$ compression ratio. 
By preserving the richness of embeddings while reducing memory usage, PQ is integral to our framework’s high scalability and its ability to perform 3D perception tasks, such as open-vocabulary 3D object localization, 3D object selection, and 3D semantic segmentation. Our contributions are summarized as follows:
\begin{itemize}
    \item We propose \nickname, direct registration and referencing of language-aligned features in 3D Gaussians, bypassing intermediate rendering and preserving feature accuracy.
    \item We introduce the PQ encoding method for compact feature representation, reducing memory usage while maintaining essential 3D feature properties.
    \item We present a novel evaluation protocol to assess accuracy of 3D localization and segmentation for 3D Gaussians, with pseudo-labeling methods and volume-aware metrics.
\end{itemize}

%% file: figures/1-teaser/teaser.tex
\begin{figure}[t!]
    \centering
    \includegraphics[width=1.0\linewidth]{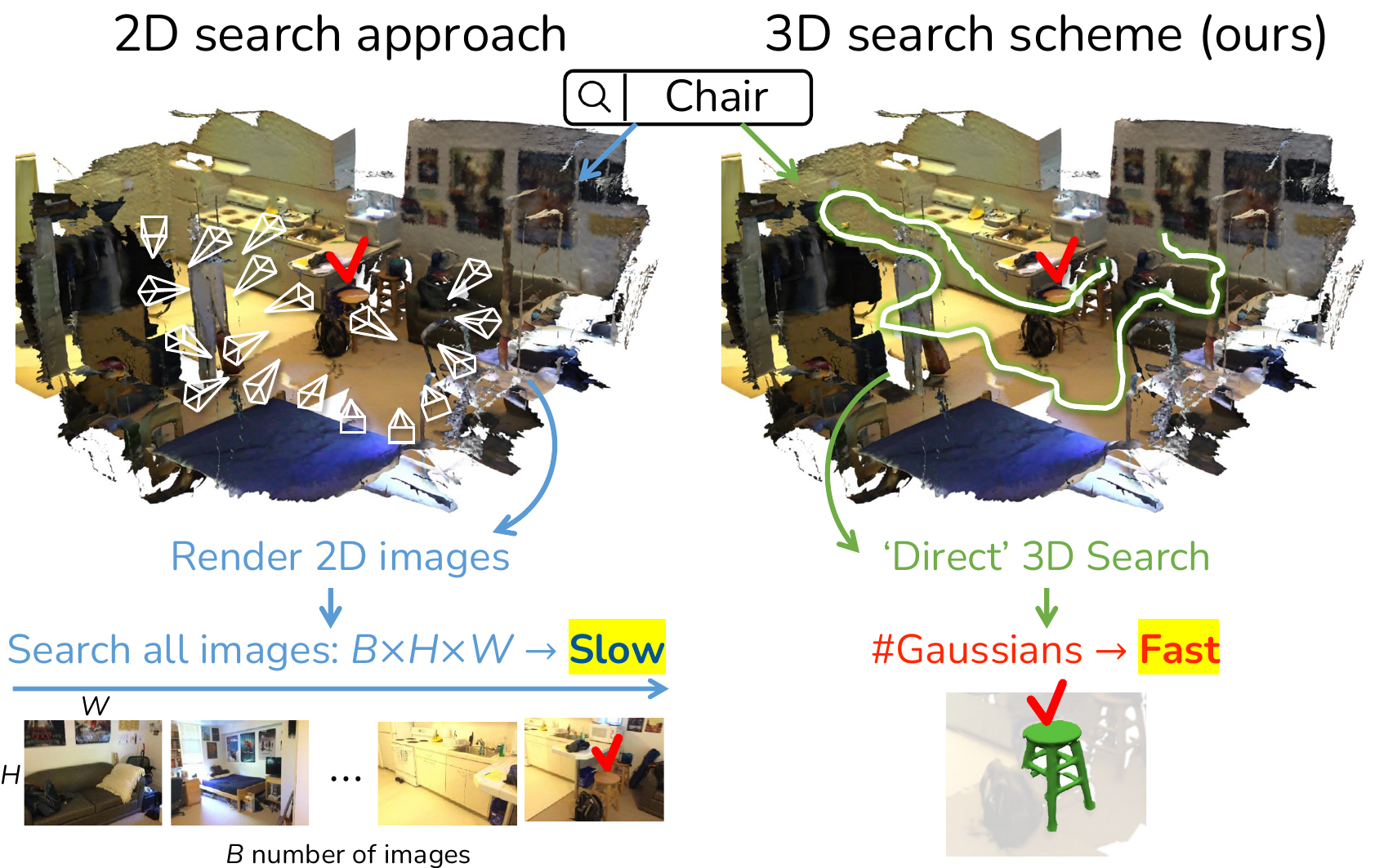}\vspace{2mm}\\
    \resizebox{\linewidth}{!}{
        \begin{tabular}{lccccc}
            \toprule
                                              & Search domain & Per-scene opt. &  Feature distill. & Search & DB size \\
            \midrule
            LERF~\cite{lerf}                  & 2D              & required             & $\sim 24h$        & slow & large               \\ 
            % \midrule
            LangSplat~\cite{langsplat}        & 2D     & required            & $\sim 4h$         & slow & large \\
            LEGaussians~\cite{legaussian}        & 2D      & required            & $\sim 4h$         & slow & large \\
            \midrule
            OpenGaussian~\cite{open_gaussian} & \textcolor{SeaGreen}{3D}      &   required                     &     $\sim 1h$         &  
            \textcolor{SeaGreen}{fast} & \textcolor{SeaGreen}{small}                                                             \\
            \textbf{\nickname (Ours)}                  & \textcolor{SeaGreen}{3D}       & \textcolor{SeaGreen}{none}                        & \textcolor{SeaGreen}{$\sim 10m$}        &  \textcolor{SeaGreen}{fast} & 
            \textcolor{SeaGreen}{small}         \\
            \bottomrule
        \end{tabular}
    }
    \vspace{-2mm}
    \caption{
        Comparison of 2D (left) vs. our direct 3D search (right) for open-vocabulary 3D scene understanding. The 2D approach relies on multiview rendering, incurring high computational costs. Our method directly links language features to 3D Gaussians, enabling efficient and complete spatial coverage. The table highlights \nickname’s superior efficiency over related methods.
    }
    \label{fig:teaser}
    \vspace{-4mm}
\end{figure}

%% file: section/2_related_works.tex
\section{Related Work and Motivation}
\label{sec:related_works}

\paragraph{Language-based 3D scene understanding}
Open-set 3D scene understanding has seen considerable advancements, with a focus on methods that leverage language knowledge into 3D representation such as point clouds, neural radiance fields (NeRF)~\cite{nerf}, and Gaussian Splatting~\cite{3dgs} for 3D comprehension. Point-based methods~\cite{regionplc,ov3d,open_scene,concept_fusion, ding2023pla, liu2023partslip, zhang2023clip} in open-vocabulary settings process point cloud data trained from language embeddings~\cite{clip,lseg} for open-set categories. 

NeRF-based approaches~\cite{lerf,liu2023weakly,kobayashi2022decomposing,opennerf,lerftogo2023} leverage semantic embeddings from 2D foundation models, such as CLIP~\cite{clip}, LSeg~\cite{lseg} and DINO~\cite{dino} for open-vocabulary understanding. While the rendering process enhances 2D perception tasks, the implicit nature of NeRF constrains the holistic understanding of 3D structures and dominantly provides `rendered' feature maps. 

3D Gaussian Splatting (3DGS)~\cite{3dgs} has emerged as a promising rendering method, as well as a novel representation for open-vocabulary 3D scene understanding. 
Since this research is the close related work with our work, we first elucidate the preliminary of 3DGS, followed by focusing on language embedded 3DGS as follows.

\input{figures/3-list/motivation}

%% file: figures/3-list/motivation.tex
\begin{figure}[t!]
    \centering
    \includegraphics[width=\linewidth]{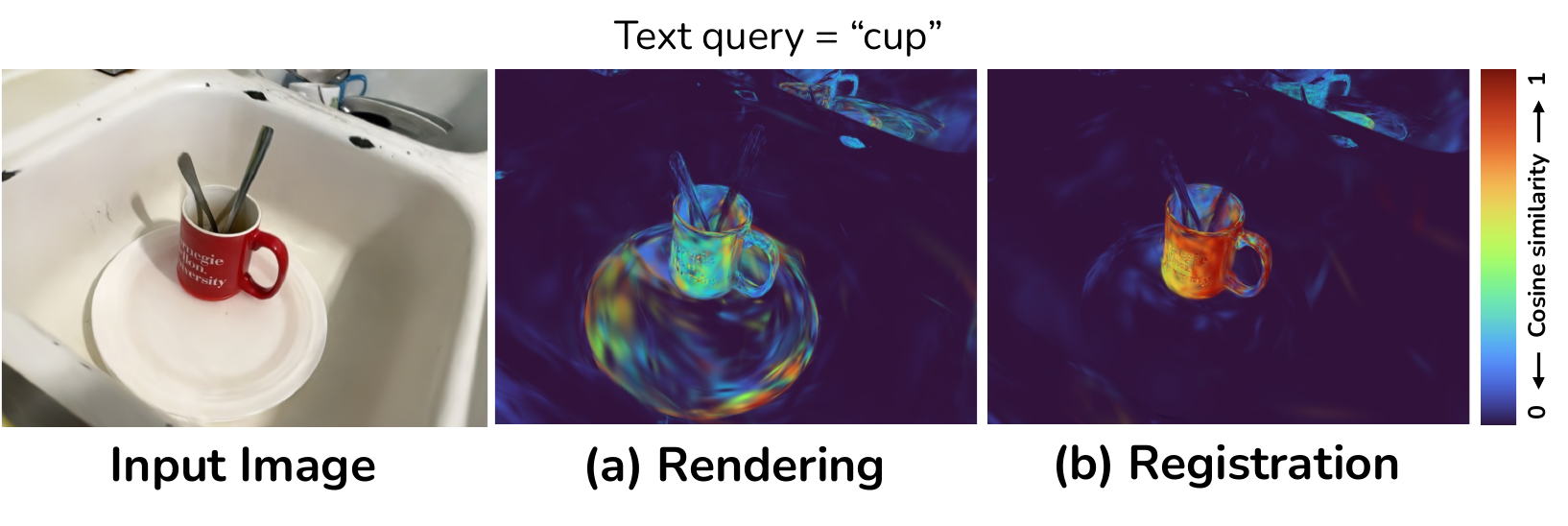}
    \vspace{-6mm}
    \caption{
        Visualization of discrepancy in rendered 2D features and 3D features. Color indicates a cosine similarity score between query features from a text query and either (a) 3D features distilled by 2D rendering~\cite{langsplat}
        or (b) directly registered 3D features. 
    }
    \label{fig:motivation}
    \vspace{-4mm}
\end{figure}

%% file: figures/2-overview/2-overview.tex
\begin{figure*}[t!]
    \centering
    \includegraphics[width=\linewidth]{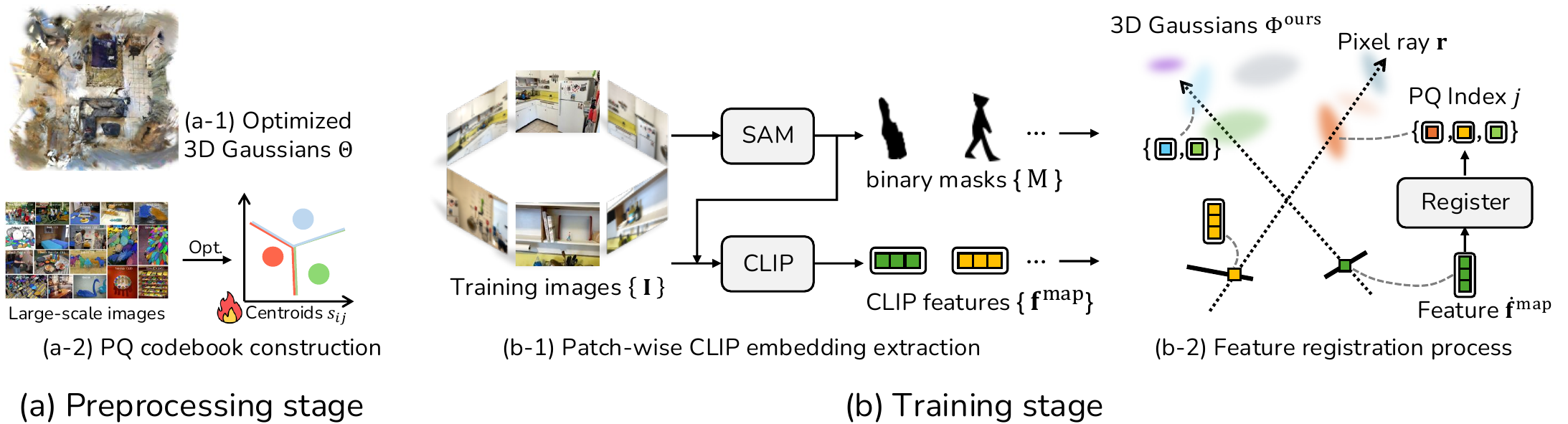}
    \vspace{-5mm}
    \caption{
        Overview of \nickname. (a) In the preprocessing stage, we compute optimized 3D Gaussians~\cite{3dgs} and Product Quantization (PQ) codebook construction. (b) During training, we extract CLIP embeddings from given images~${\{\mathbf{I}\}}$,
        and then proceed feature registration process (\cref{subsec:4_1}). Finally, we obtain 3D Gaussians~$\Phi^\text{ours}$ with PQ indices $\{ j \}$ (\cref{subsec:4_2}).
    }
    \label{fig:overview}
\end{figure*}

%% file: section/3_preliminary.tex
\label{sec:preliminary}

\paragraph{Preliminary of 3D Gaussian Splatting}
3DGS~\cite{3dgs} encodes appearance and geometry of the target scene into the 3D Gaussian representation. Each 3D primitive representation is expressed as a 3D Gaussian distribution having mean~${\bmu}=[x_\mu, y_\mu, z_\mu]^\top$ for 3D position and covariance matrix~$\Sigma_\text{3D} \in \mathbb{R}^{3\times 3}$ for 3D volume, as well as the opacity value~$\alpha$ and the color~$\mathbf{c}$. In particular, the covariance matrix is decomposed into the scale matrix~$S \in \mathbb{R}^{3\times3}$ and the rotation matrix~$R \in SO(3)$,
$\Sigma_\text{3D} = R S S^\top R^\top$.
In brief, $N$ numbers of 3D Gaussians can be parametrized as~$\Theta = \{ \btheta_i \}_{i=1}^N = \{ \bmu_i, S_i, R_i, \alpha_i, \mathbf{c}_i \}_{i=1}^N$.
3D Gaussians~$\Theta$ are used to render a 2D pixel color~$\hat{\mathbf{c}}$ computed as:
\begin{equation}
\label{eq:vol_render_in_3dgs}
    \hat{\mathbf{c}}(\theta) {=} \sum\nolimits_{i=1}^{N} \!\!T_i\tilde{\alpha_i}\mathbf{c}_i,~\textrm{s.t.}~~\tilde{\alpha_i} {=} \alpha_i \mathrm{exp}\left({ - \tfrac{1}{2}  \mathbf{d}^\top \Sigma_\text{2D}^{-1} \mathbf{d} }\right) ,
\end{equation}
$T_i$ is a transmittance, $\tilde{\alpha_i}$ is an effective opacity value computed from the Gaussian's opacity 
$\alpha$, the pixel distance~$\mathbf{d} \in \mathbb{R}^{2\times 1}$ from the target pixel to the projected center location of the Gaussian in pixel, and $\Sigma_\text{2D}$ is the 2D covariance matrix in the image domain obtained from the splatting algorithm~\cite{3dgs,ewa}. The 3D Gaussian parameters~$\Theta$ of a scene are optimized by minimizing the rendering loss between the input image color $\mathbf{{c}}$ and the rendered color $\mathbf{\hat{c}(\theta)}$ in \Eref{eq:vol_render_in_3dgs} as $\mathop{\arg\min}_{\theta} \| \mathbf{{c}} - \mathbf{\hat{c}}(\theta) \|_F^2$.

\paragraph{Language embedded 3D Gaussian Splatting}
The basic idea of the language embedded Gaussian representation~\cite{langsplat, legaussian, zhou2024feature, semantic_gaussians, fastlgs, fmgs, gaussian_grasper, rethinking_open_vocab} is to replace the color rendering to language embedding rendering. 
Language embedded 3D Gaussians are parameterized as $\Phi = \{ \theta_i, \mathbf{\tilde f}_i \}_{i=1}^N = \{ \bmu_i, S_i, R_i, \alpha_i, \mathbf{c}_i, \mathbf{\tilde f}_i \}_{i=1}^N$, where $\mathbf{\tilde f}_i$ denotes Gaussian-registered language embeddings across $N$ numbers 3D Gaussians which will be discussed soon. Then, analogous to the color rendering \cref{eq:vol_render_in_3dgs}, the language embedding rendering is expressed as:
\begin{equation}
\label{eq:feat_render_in_3dgs}
    \mathbf{\hat f} = \sum\nolimits_{i=1}^{N} T_i\tilde{\alpha_i}\mathbf{\tilde f}_i, 
\end{equation}
where $\mathbf{\hat f}$ denotes a rendered language embedding.
Likewise, the Gaussian-registered language embeddings $\{\mathbf{\tilde f}\}$ are optimized by minimizing the rendering loss between the 2D language embedding~$\mathbf{f}$ extracted from an input image and a rendered language embedding map $\hat{\mathbf{f}} $ as $\mathop{\arg\min}_{\{\mathbf{\tilde f}\}} \| \mathbf{{f}} - \mathbf{\hat{f}} \|_F^2$ at each corresponding pixel. 
This can be regarded as distilling vision language models into Gaussian-registered language embedding~$\mathbf{\tilde{f}}$ through volume rendering~\cref{eq:feat_render_in_3dgs}.
The Gaussian-registered language embeddings are separately trained after pre-training and fixing the pre-trained 3DGS $\Theta$ for a scene.
The language embeddings to be distilled are typically obtained from CLIP~\cite{clip}.
Since storing 32-bit 512-D CLIP features~$\mathbf{f}$ in every 3D Gaussians is memory-expensive, one can use a compressed feature per scene depending on the needs~\cite{fastlgs,legaussian, zhou2024feature,fmgs, gaussian_grasper}.

\paragraph{Motivation}
Such language-embedded radiance fields provide useful representation and language interfaces for many practical and crucial applications.
While most of existing works focus on the training efficiency, the complexity in inference time has barely been discussed.
Considering a scenario to text-query a 3D location of the language-embedded Gaussians, \ie, 3D localization, the aforementioned methods first require rendering a 2D language embedding map at each specific camera pose.
We cannot directly retrieve over the distributed embeddings $\{\mathbf{\tilde f}_i\}$ in 3D Gaussians, because the embeddings do not carry language information directly, but their weighted summed (rendered) features $\mathbf{\hat{f}}$ do.
This issue becomes even severer with compressed features as in~\cite{langsplat}: their decompression decoders are not designed for and incompatible with directly applying to the distributed compressed language embeddings in each 3D Gaussian, yielding degenerated CLIP decoding (refer to~\cref{fig:motivation}).

This introduces multiple challenges and hassles.
First, it is challenging to find the best or proper camera rendering views that contain the object to find.
One may attempt to pre-compute the minimal number of cameras and their camera poses that cover all the 3D Gaussians in a scene with proper resolutions, similarly by point-based approach~\cite{openins3d}. However, this is a well-known set covering problem~\cite{garey1982computers} with constraints which is known to be an NP-hard problem.

Second, even with pre-computed rendered views, the retrieval complexity over the rendered images remains substantial~\cite{guedon2023macarons}.
Suppose a scene consisting of one million Gaussians, but just a \emph{single} rendered language embedding map in pixel domain already has nearly a million pixels; thus, we need a dedicated system to efficiently retrieve over all the views.
Third, since the retrieval is conducted in the 2D space, to find a 3D location, we need a separate mechanism to lift the localization to the 3D space, \ie, increasing the system complexity.
In addition, 32-bit floating 512-Dimension CLIP features for millions of Gaussian are memory intensive, which is often not manageable. To reduce this burden, the existing methods~\cite{open_gaussian} apply compressions with per-scene optimized codebooks, which hinders extension or generalization to other scenes.

To overcome these, we propose a training-free algorithm for the direct allocation of language embeddings to 3D Gaussians, allowing efficient computation and interaction within the 3D space. As a concurrent work, OpenGaussian~\cite{open_gaussian} tackles a similar challenge with our work, but still requires per-scene codebook construction~\cref{fig:teaser}.

%% file: figures/4-pq/4-pq.tex
\begin{figure*}[t!]
    \centering
    \includegraphics[width=0.99\linewidth]{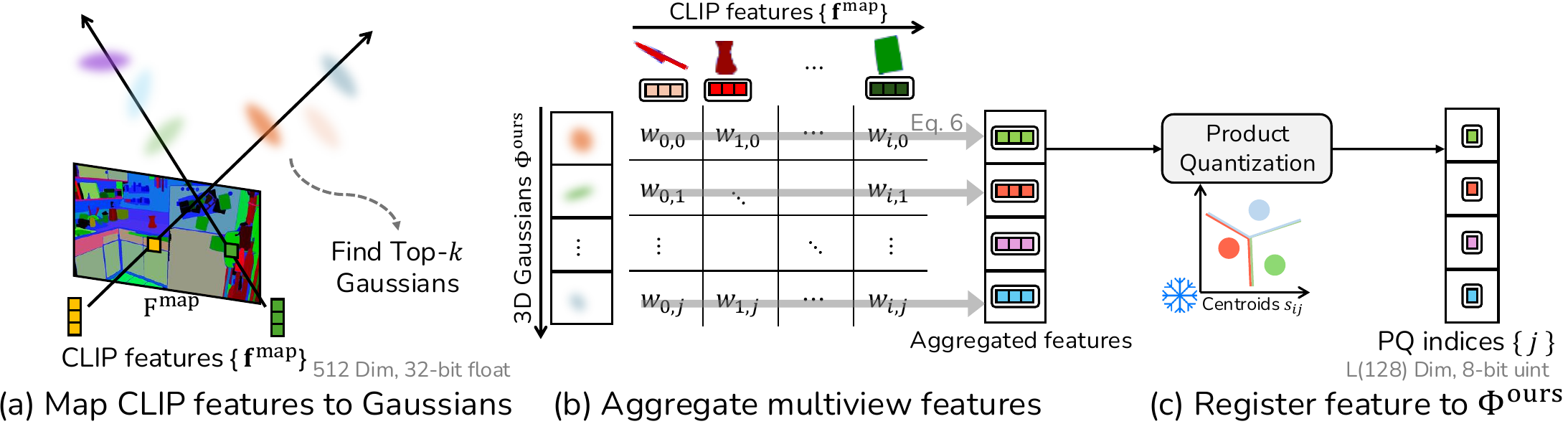}
    % \vspace{-2mm}
    \caption{
        Feature registration process in \nickname.
        (a) We first map per-pixel CLIP embeddings $\{ \mathbf{f}^\text{map} \}$ to Gaussians. Here, we only map dominant $k$ Gaussians along pixel ray~$r$, named Top-$k$ Gaussians. (b) After collecting embeddings, we compute aggregated features (\cref{eq:weighted-averaging}). (c) Finally, we re-use PQ to obtain the PQ indices $j$ of aggregated features and update Gaussian parameters~$\Phi^\text{ours}$.
    }
    \label{fig:inference}
\end{figure*}

%% file: section/4_0_method.tex
\section{\nickname}
\label{sec:methodology}

This section provides details of our method. 
We first explain how we directly register CLIP embeddings into Gaussian-registered language embeddings,~\cref{subsec:4_1}. 
Then, we introduce Product Quantization (PQ) into our framework to efficiently store Gaussian-registered language embeddings,~\cref{subsec:4_2}.
Lastly, we describe the inference stage for text query-based 3D Gaussian localization,~\cref{subsec:4_3}.

%% file: section/4_1.tex
\subsection{Feature registration process}
\label{subsec:4_1}

Our goal is to reconstruct a language embedded 3D space represented by 3D Gaussians~$\Phi$, which we can directly interact in 3D space without feature rendering~\cref{eq:feat_render_in_3dgs}. For that, following LangSplat~\cite{langsplat}, we begin by extracting per-pixel CLIP embedding maps $\mathbf{F}^\text{map} \in \mathbb{R}^{D\times H \times W}$ from training images of the target scenes, where $D$ is the dimension of CLIP embeddings, $H$ and $W$ are the height and width of the training images. Given training images, we extracts a dictionary of binary masks and language embeddings extracted from the images as: $\mathcal{F}^\text{map} = \{\mathbf{M}_j : \mathbf{f}^\text{map}_j \ |\ j=1,...,M\}$, where $\mathbf{M}_j\in \mathbb{R}^{H \times W}$ is a binary mask extracted using SAM~\cite{sam} and $\mathbf{f}^\text{map}_j \in \mathbb{R}^{D}$ is a corresponding CLIP embedding 
from
a cropped image with $\mathbf{M}_j$.
Each mask $\mathbf{M}_j$ belongs to an image, and the masks are not overlapped to each other. 
With this dictionary, a CLIP embedding map~$\mathbf{F}^\text{map}(\mathbf{I}, \mathbf{r})$ at a pixel $\mathbf{r}$ in a training image $\mathbf{I}$ is computed as:
\begin{equation}
    \mathbf{F}^\text{map}(\mathbf{I}, \mathbf{r}) = \sum\nolimits_{j=1}^M\mathbf{M}_j(\mathbf{I}, \mathbf{r})\cdot\mathbf{f}^\text{map}_j,
\end{equation}
where $\mathbf{M}_j(\mathbf{I}, \mathbf{r}) \in \{0,1\}$ indicates whether the mask $\mathbf{M}_j$ contains the pixel $\mathbf{r}$ in the image $\mathbf{I}$. 
Using $\mathbf{F}^\text{map}$, we reconstruct language embedded 3D Gaussians via a novel feature registration process as visualized in~\cref{fig:overview}. 

During the feature registration process, our algorithm iterates through training images of the scene. Using projection relation, we link 3D Gaussians~$\Phi$ to CLIP embeddings. Each Gaussian can link to multiple CLIP embeddings derived from different images. Then we aggregate collected embeddings to a single embedding to be assigned to each Gaussian.
To ensure a consistent aggregation of the embeddings from multi-view images, we first compute a weight $w_i(\mathbf{I}, \mathbf{r})$ representing the contribution of $\theta_i$ to construct each pixel $\mathbf{r}$ in a training image $\mathbf{I}$.
The weights are computed with the volume rendering equation \Eref{eq:vol_render_in_3dgs} as:
\begin{equation}
    w_i(\mathbf{I}, \mathbf{r}) = T_i(\mathbf{I}, \mathbf{r})\cdot\tilde{\alpha_i}(\mathbf{I}, \mathbf{r}),
    \label{eq:weight definition}
\end{equation}
where $T_i(\mathbf{I}, \mathbf{r})$ and $\tilde{\alpha_i}(\mathbf{I}, \mathbf{r})$ are the transmittance and the effective opacity value of $\theta_i$ for a pixel $\mathbf{r}$ in an image $\mathbf{I}$, stated in~\cref{eq:vol_render_in_3dgs}. With the per-pixel weights, we calculate $w_{ij}$ representing a weight between each Gaussian $\theta_i$ and corresponding language embedding maps $\mathbf{f}_j^\text{map}$, which is for aggregating CLIP embeddings from $\mathbf{F}^\text{map}$ and register the embedding to each Gaussian. The weights are computed as:
\begin{equation}
    w_{ij} = \sum\nolimits_{\mathbf{I}\in \mathcal{I}}\sum\nolimits_{\mathbf{r} \in \mathbf{I}} \mathbf{M}_j(\mathbf{I},\mathbf{r})\cdot w_i(\mathbf{I},\mathbf{r}),
\end{equation}
where $\mathcal{I}$ is the set of the training images. In this iterative process, we aggregate weights only for Top-$k$ Gaussians with the highest weights $w_i(\mathbf{I}, \mathbf{r})$, along the ray of each pixel ray~$\mathbf{r}$ (see \Fref{fig:inference}). After aggregation, we prune the Gaussians which are not assigned any weight, \ie, $\sum_{j=1}^M w_{ij}=0$.
This summation aggregates weights between Gaussians and the CLIP embeddings by linking per-pixel weights $w_i(\mathbf{I},\mathbf{r})$ of each Gaussian to its corresponding CLIP embeddings. 
With the obtained weights, we register an aggregated feature $\dot{\mathbf{f}}_i$ to each Gaussian with weighted-averaging as:
\begin{equation}
    \label{eq:weighted-averaging}
    \begin{gathered}
            \dot{\mathbf{f}}_i = \mathbf{f}_i / ||\mathbf{f}_i||_2, \,\, \textrm{where}\quad \mathbf{f}_i = \sum\nolimits_{j=1}^{M} \tfrac{w_{ij}}{\sum^M_{k=1} w_{ik}} \mathbf{f}^\text{map}_j.
    \end{gathered}
\end{equation}
This process enables 3D-aware feature registration to be consistent across various viewpoints, by aggregating features in the original high-dimensional feature space. The proposed process can be interpreted as an inverse volume rendering without gradient-based optimization, which enables our method to be faster than the prior methods requiring per-scene gradient-based optimization~\cite{langsplat, open_scene,legaussian} for feature registration in 3D space.

%% file: tables/Object_selection.tex
\begin{table*}[t]
    \centering
    
    \label{table:Object_selection}
    \resizebox{1.0\linewidth}{!}{
        \begin{tabular}{c|ccccc|ccccc}
        \toprule
        \multirow{2}{*}{Methods} & \multicolumn{5}{c|}{mIoU} & \multicolumn{5}{c}{mAcc @ 0.25} \\
           & waldo\_kitchen             & ramen               & figurines                 & teatime        & Mean & \multicolumn{1}{c}{waldo\_kitchen} & \multicolumn{1}{c}{ramen} & \multicolumn{1}{c}{figurines} & \multicolumn{1}{c}{teatime} & Mean \\ 
       \midrule
        LangSplat-m~\cite{langsplat}       & 8.29          & 6.11          & 8.33              & 16.58      & 9.83
                        & 13.64          & \underline{14.08}          & 8.93              & 27.12      & 15.94    \\
        OpenGaussian~\cite{open_gaussian}    & 34.60 	   & 23.87	        & \textbf{59.33}	            & 54.44	     & 43.06				
                        & \underline{50.00} 	   & \textbf{35.21}	        & \underline{80.36}	            & 72.88	     & 59.61     \\
        Ours (Top-10)  & 37.05 	   & 24.33	        & \underline{54.42}	            & \textbf{57.35}	     & \underline{43.29}
                        & \textbf{63.64} 	   & \textbf{35.21}	        & \underline{80.36}	            & \textbf{77.97}	     & \underline{64.30}     \\
        Ours (Top-20)  & \underline{38.33} 	   & \underline{24.58}	        & 53.94	            & 56.19	     & 43.26
                        & \textbf{63.64} 	   & \textbf{35.21}	        & \textbf{82.14}	            & \underline{76.27}	     & \textbf{64.32}     \\
        Ours (Top-40)  & \textbf{39.07}& \textbf{24.70}	        & 53.36	            & \underline{57.20}	     & \textbf{43.58}
                        & \textbf{63.64} 	   & \textbf{35.21}	        & \underline{80.36}	            & \underline{76.27}	     & 63.87     \\
        \bottomrule
        \end{tabular}

    \vspace{-4mm}
    }
    \caption{
        3D object selection results on the LeRF-OVS dataset~\cite{lerf}.
        To measure 3D object selection performance, we calculate 2D segmentation accuracy on rendering of selected 3D Gaussians. 
        Note that our model does not require per-scene optimization, demonstrating its robustness across diverse scenes. 
        \textbf{Bold} and \underline{Underline} stand for first and second best performance.
    }    \label{table:obj_sel}
\end{table*}

%% file: figures/5-obj_sel/5-obj_sel.tex
\begin{figure*}[t!]
    \centering
    \includegraphics[width=\linewidth]{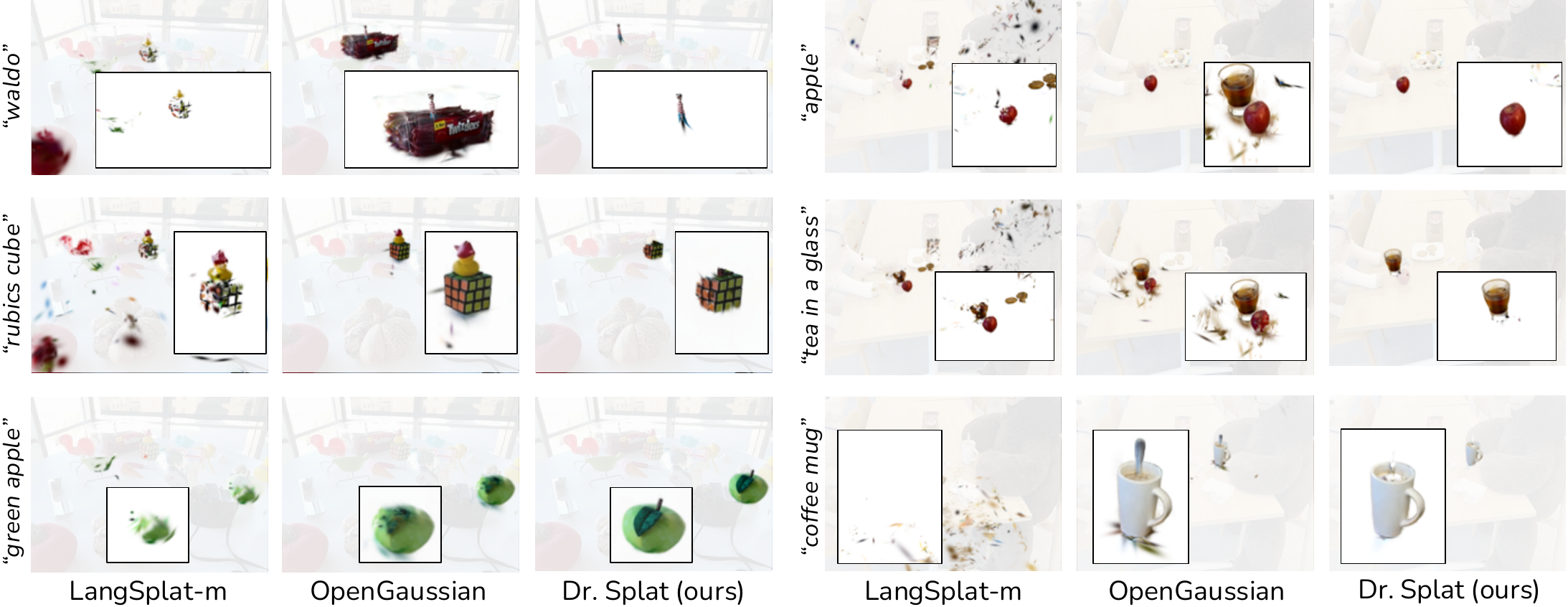}
    \vspace{-6mm}
    \caption{
        Qualitative results of the object selection on the LeRF-OVS dataset~\cite{lerf}.
        We visualize rendering of selected 3D Gaussians for LangSplat~\cite{langsplat}, OpenGaussian~\cite{open_gaussian}, and ours. For LangSplat, activations are often distributed randomly, fail to localize the target.  OpenGaussian often struggles to distinguish closely situated objects. In contrast, our model shows activations precisely limited to the queried object regions, effectively localizing only the relevant areas.
    }
    \label{fig:5_obj_sel}
\end{figure*}

%% file: section/4_2.tex
\subsection{Product-Quantized CLIP embeddings}
\label{subsec:4_2}

Memory efficiency is a challenge in 3D scene representations, especially when associating Gaussians with high-dimensional feature vectors. LangSplat~\cite{langsplat} addresses this by introducing an encoder-decoder network, while LeGaussian~\cite{legaussian} and OpenGaussian~\cite{open_gaussian} utilize codebook construction. However, these approaches introduce additional per-scene computational costs for scene-specific parameter tuning of neural networks or codebooks (see \cref{fig:teaser}). In contrast, we propose to use Product Quantization (PQ) on a large-scale image dataset, eliminating per-scene training.

%% file: tables/3d_loc_and_seg.tex
\begin{table*}[!t]
    \centering
    \begin{subtable}[t]{.47\linewidth}
        \centering
        \resizebox{1.0\linewidth}{!}{ 
            \begin{tabular}{l|c|ccc}
                \toprule
                & 3D   & \multicolumn{3}{c}{19 classes} \\
                \multicolumn{1}{c|}{} & mIoU & IoU $>$ 0.15 & IoU $>$ 0.3 & IoU $>$ 0.45 \\
                \midrule
                LangSplat-m~\cite{langsplat} & 8.0  & 17.1 & 7.8 & 2.9 \\
                LEGaussians-m~\cite{legaussian} & 9.5  & 19.1 & 8.9 & 7.3 \\
                OpenGaussian~\cite{open_gaussian} & \underline{25.2} & \underline{59.5} & \underline{38.0} & 18.3 \\
                Ours (Top-20) & 25.0 & \textbf{60.7} & \textbf{40.3} & \underline{20.0} \\
                Ours (Top-40) & \textbf{25.4}  & \textbf{60.7} & \textbf{40.3} & \textbf{25.6} \\
                \bottomrule
            \end{tabular}
        }
        \caption{
            3D object localization task.
        }
        \label{table:3d_loc}    
    \end{subtable}
    \hspace{1mm}
    \begin{subtable}[t]{.495\linewidth}
        \centering
        \resizebox{1.0\linewidth}{!}{
            \begin{tabular}{l|cccccc}
                \toprule
                \multicolumn{1}{c|}{\multirow{2}{*}{}} & \multicolumn{2}{c}{19 classes} & \multicolumn{2}{c}{15 classes} & \multicolumn{2}{c}{10 classes} \\
                \multicolumn{1}{c|}{} &  mIoU & mAcc. & mIoU & mAcc. & mIoU & mAcc. \\
                \midrule
                LangSplat-m~\cite{langsplat} &  2.0 & 9.2 & 4.9 &  14.6 & 8.0 & 23.9 \\
                LEGaussians-m~\cite{legaussian} &  1.6 & 7.9 & 4.6 &  16.1 & 7.7 & 24.9 \\
                OpenGaussian~\cite{open_gaussian} &  \textbf{30.1} & \underline{46.5} & \underline{38.1} &  \underline{56.8} & \underline{49.7} & \underline{71.4} \\
                Ours (Top-20) &  28.0 & 44.6 & \textbf{38.2} &  \textbf{60.4} & 47.2 & 68.9 \\
                Ours (Top-40) &  \underline{29.6} & \textbf{47.7} & \textbf{38.2} &  \textbf{60.4} & \textbf{50.2} & \textbf{73.5} \\
                \bottomrule
            \end{tabular}
        }
    \caption{
        Open-vocabulary 3D semantic segmentation task.
    } 
    \label{table:3d_seg}
    \end{subtable}
    \caption{
        Quantitative comparison in the ScanNet dataset~\cite{dai2017scannet}.
        Left: Localization prediction is defined as 3D regions with a text similarity score above threshold. 
        Right: We assign segmentation labels by finding max activations among all classes. 
        Note that \textbf{Bold} and \underline{Underline} stand for first and second best performance, respectively.
    }
    \label{table:3d_loc_and_seg}
\end{table*}

%% file: figures/7-weight_verification/7_weight_verification.tex
\begin{figure}[t!]
    \centering
    \includegraphics[width=\linewidth]{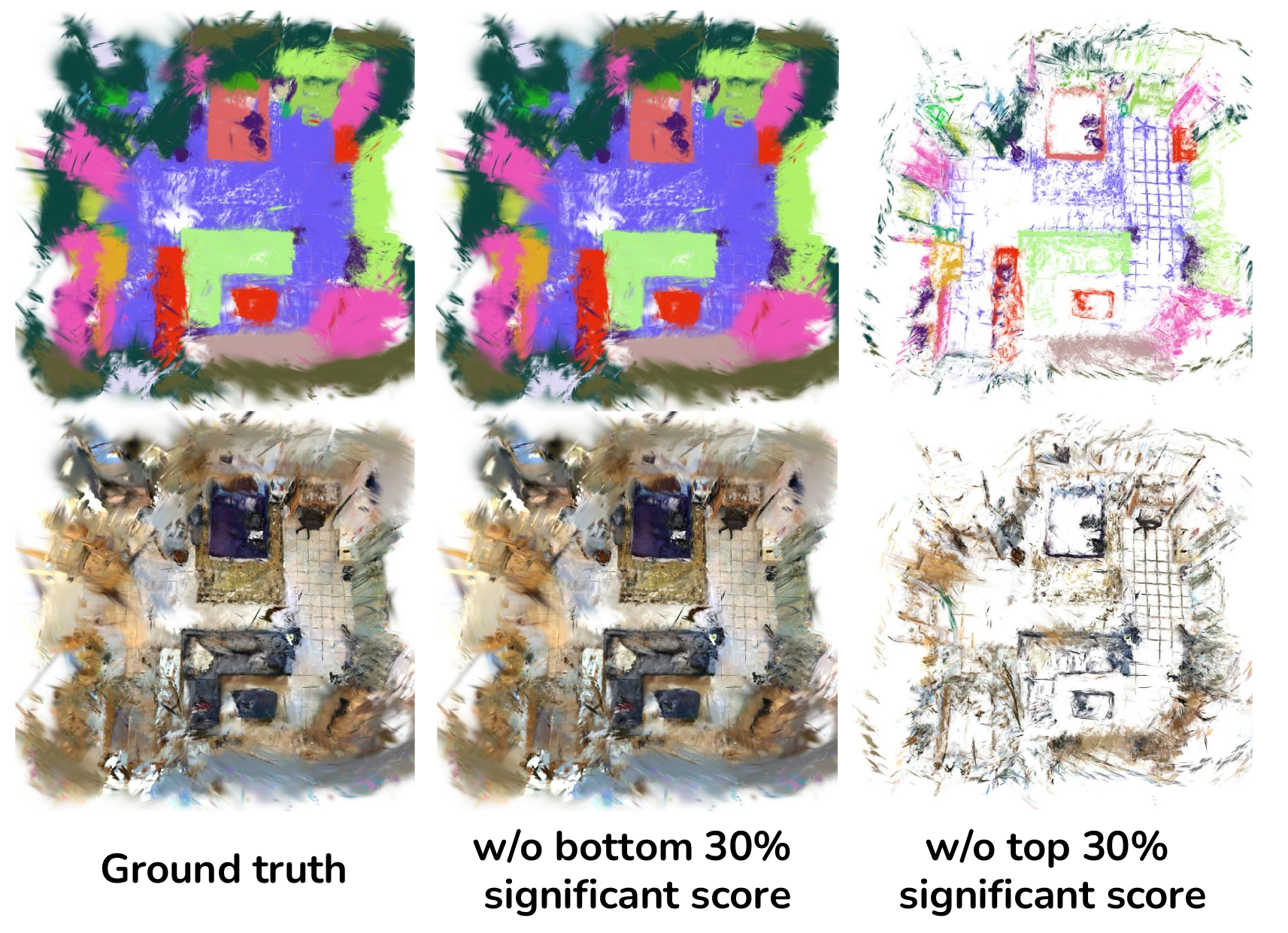}
    \vspace{-6mm}
    \caption{
        Limitations of point-based IoU measurement.
        This figure shows the effect of removing the top and bottom 30\% of Gaussians according to the proposed significant score, implying 
        % demonstrating
        that volume differences significantly impact 3D accuracy. The results highlight the need for the proposed IoU metric for 3D Gaussians.
    }
    \label{fig:significant_score}
    \vspace{-5mm}
\end{figure}

%% file: figures/6-loc_and_seg/6-loc.tex
\begin{figure*}[t!]
    \centering
    \includegraphics[width=1\linewidth]{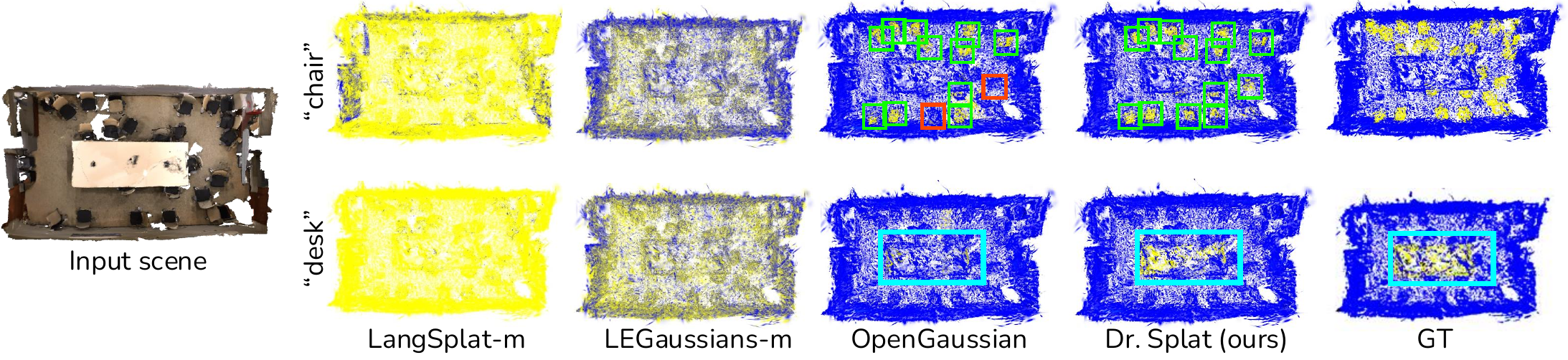}
    \caption{
        Qualitative results of 3D object localization. We visualize 3D localization activations (yellow) for “chair” and “desk” in the ScanNet dataset, comparing our method with others. It turns out that LangSplat-m and LEGaussians-m fail to localize objects accurately, while OpenGaussian struggles with object correspondence. Our model delivers precise and consistent localization across diverse queries.
    }
    \label{fig:6_seg}
\end{figure*}

%% file: section/4_3.tex
\paragraph{Product Quantization}
PQ~\cite{product_quantization} is a widely used technique for efficient embedding compression, particularly valuable in large-scale applications.
The PQ process begins by dividing the original $D$-dimensional feature vector $\mathbf{v}$ into $L$ sub-vectors: $\mathbf{v} = [\mathbf{v}_1, \mathbf{v}_2, \ldots, \mathbf{v}_L]$. 
Each sub-vector $\mathbf{v}_i$ is then independently quantized to a predefined number of centroids $s_{ij}$ in a predefined codebook $S_i$ for that sub-vector. These centroids are learned via clustering, creating a codebook for each subspace. Once the centroids are established, each sub-vector is replaced by the index of the nearest centroid in its respective codebook. 
The centroid indices~$j_i = [j_{i1}, j_{i2}, \ldots, j_{iL}]$ are optimized by minimizing $\mathop{\arg\min}_{k} \lVert \mathbf{v}_i - s_{ik} \rVert$ to quantize a given vector $\mathbf{v}_i$ where $j_{ik}$ is an 8-bit unsigned integer. 

Then, we can measure the distance between the query and data by adding distances between coarse centroids. Once the distances between centroids are computed as a lookup table, the computation shifts to simple indexing, which reduces the search complexity from $\mathcal{O}(D)$ to $\mathcal{O}(1)$ for a $D$ dimension sample. This approach notably reduces computational complexity, making it suitable for large-scale search.

In our setup for language-based 3D scene understanding,
we build PQ centroids based on CLIP embeddings using a large-scale image dataset, the LVIS dataset~\cite{lvis_dataset}, that contains over $1.2$M instances covering various long-tail classes and ground truth segmentation. We extract instance patches from images and collect patch-wise CLIP embeddings. After we build this CLIP embedding database, we proceed with the construction of the centroid codebook for our PQ. Once PQ is trained, any query embedding can be approximated by assigning the closest centroid for each subvector. This is a one-time procedure; once we determine the codebook, we can use it for any scene generally.
In our setup, each embedding is represented as a sequence of centroid indices rather than a high-dimensional vector. Accordingly, our language embedded Gaussians are parametrized as $\Phi^\text{ours} = \{ \phi_i^\text{ours} \}_{i=1}^N = \{ \theta_i, j_i \}_{i=1}^N$. where the aggregated feature~$\dot{\mathbf{f}}_i$ are converted as a quantized feature~$\bar{\mathbf{f}}_i$ by the corresponding PQ index $j_i$.

\input{figures/6-loc_and_seg/6-2-seg}

\subsection{Text-query based 3D localization}
\label{subsec:4_3}

After training 3D Gaussians~$\Phi^\text{ours}$ with our feature registration process and PQ, we describe the details of an inference mode that facilitates direct interaction with 3DGS upon receiving input queries, such as text. 
This is related to similarity score computation between a query and sources, \ie Gaussian embeddings. Given a text, we first extract a query feature~$\mathbf{q}$ using CLIP text encoder~\cite{clip}. We reconstruct the quantized features~$\{ \bar{\mathbf{f}}_i \}_{i=1}^N$ from the stored PQ indices~$\{ j_i \}_{i=1}^N$. Then, we compute a cosine similarity score between the query feature~$\mathbf{q}$ and all quantized features.

Despite its simplicity, solely relying on the cosine similarity may result in diminished discriminability across certain similarity scores.
 
To address this limitation, we incorporate a re-ranking process based on relative activation with respect to the canonical feature. For this process, we adopt the relevancy scoring method proposed in LeRF~\cite{lerf}, which enables more precise similarity analysis for a query.
Specifically, each rendered language embedding, $\featmap$ and a text query feature~$\mathbf{q}$, yield a relevance score determined by, 
    $\mathop{\min}_i \frac{
        \exp(\featmap \cdot \mathbf{q} )
    }{
        \exp(\featmap \cdot \mathbf{q} ) + \exp(\featmap \cdot \mathbf{f}^\text{canon, i})
    },$
where $(\cdot)$ is an element-wise dot product operator and $\mathbf{f}^\text{canon,i}$ indicates CLIP embeddings of a designated canonical term selected from a set of ``object,'' ``things,'' ``stuff,'' and ``texture''. Then, we sample 3D Gaussians based on the relevance score for downstream tasks.

%% file: figures/6-loc_and_seg/6-2-seg.tex
\begin{figure*}[t!]
    \centering
    \includegraphics[width=\linewidth]{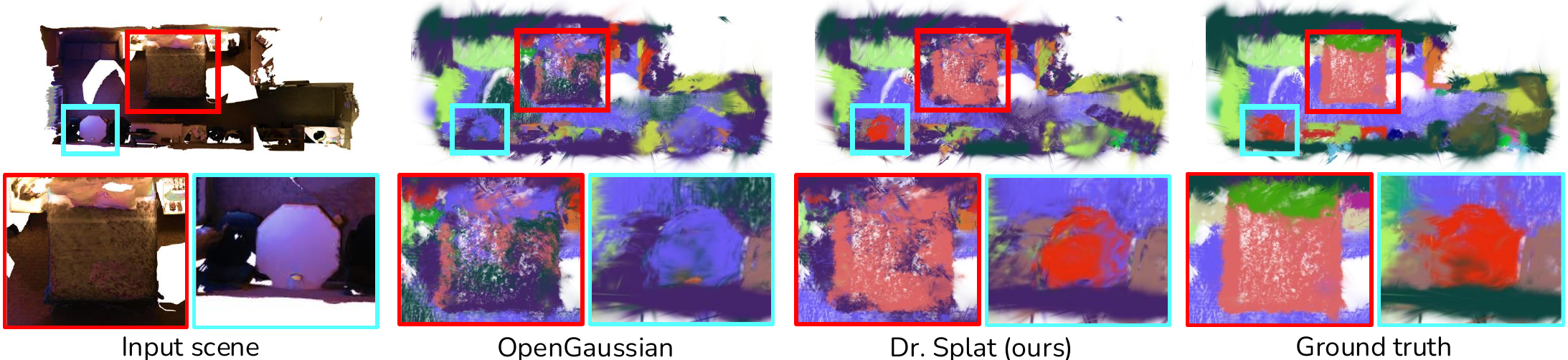}
    \vspace{-2mm}
    \caption{
        Visualization of open-vocabulary 3D semantic segmentation on the ScanNet dataset~\cite{dai2017scannet}.
        We visualize 3D Gaussian splat-based semantic segmentation using language features allocation of  OpenGaussian~\cite{open_gaussian} and \nickname (ours) model on the same RGB-pretrained 3DGS. Note that, not specifically designed for segmentation, it achieves high performance as a result of language-based Gaussian updates.
    }
    \label{fig:6-2-seg}
\end{figure*}

%% file: section/5_0_experiment.tex
\section{Experiments} 
\label{sec:experiment}

\noindentbold{Dataset}
We use two datasets to evaluate the 3D scene understanding performance. For the 3D object selection task (\Sref{subsec:5_1}), we use the LERF~\cite{lerf} dataset annotated by LangSplat~\cite{langsplat}, which consists of several multi-view images of 3D scenes containing long-tail objects and includes ground truth 2D ground truth annotations for texture queries. For 3D object localization \Sref{subsec:5_2} and 3D semantic segmentation \Sref{subsec:5_3} task, we employ the ScanNet~\cite{dai2017scannet} dataset. ScanNet is a large-scale benchmark that provides data on indoor scenes, including calibrated RGBD images and 3D point clouds with ground-truth semantic labels. We randomly select eight scenes from ScanNet for the experiments.

\noindentbold{Competing methods}
The only method available for a fair comparison with our method is the concurrent work, OpenGaussian~\cite{open_gaussian}.
To study the various aspects of our method, we introduce baseline methods modified from rasterization-based ones \cite{langsplat,legaussian}, for direct 3D referring operation, denoted as LangSplat-m and LEGaussians-m.
As discussed in \Sref{sec:related_works}, without modification, global search over a whole scene is quite demanding. To ensure fair evaluation, we use the same initial 3D Gaussians being trained only using RGB inputs for all comparing methods, and freeze the Gaussians during the language feature allocation process. Also, the per-pixel CLIP~\cite{clip} embedding maps are unified for SAM-based~\cite{sam} methods~\cite{langsplat, open_gaussian} including ours. We follow the hyperparameter settings favorable to each respective paper.

%% file: section/5_1.tex
\subsection{3D object selection}
\label{subsec:5_1}

\noindentbold{Settings}
We first extract text features from an open-vocabulary text query using the CLIP model. Next, we compare text features to the 3D features embedded in each Gaussian using cosine similarity. By thresholding the similarity, we identify the 3D Gaussians that are relevant to the given text query. The selected 3D points are subsequently rendered into multi-view images using the 3DGS rasterization pipeline. 

\noindentbold{Results}
We compare our model quantitatively with 3DGS-based language-embedded models as shown in~\Tref{table:obj_sel}. 
The results demonstrate that our method performs better object selection in most scenes, showing an improvement of over 0.5 in mIoU and more than 4.5 in mAcc compared to counterpart models. Notably, the rasterization-based method, LangSplat-m, often underperforms in most scenes.

Qualitative results are shown in~\Fref{fig:5_obj_sel}. 
For LangSplat-m, the activations often shows random 3D Gaussians or fail to localize entirely (\eg, see ``coffee mug''), highlighting the limitations of rasterization-based methods and their unsuitability for 3D understanding, aligning the observation from~\cref{fig:motivation}. OpenGaussian frequently exhibits false activations with incorrect text-object pairs (\eg, ``apple'' and ``tea in a glass'') and struggles to distinguish between nearby objects (\eg, ``waldo,'' ``rubik's cube'').
This artifacts can be attributed to use of spatial clustering and limited encoder capacity.

In contrast, our model leverages general image features thanks to the general PQ, maintaining feature distinctiveness regardless of scene complexity.
Our feature registration considers the 3D geometry of the 3D Gaussians, which results in superior performance in 3D scene understanding tasks.

%% file: section/5_2.tex
\subsection{3D object localization}
\label{subsec:5_2}

\noindentbold{Settings}
Similar to the 3D object selection task, we calculate the cosine similarity between text query and 3D features embedded in each Gaussian. By thresholding the similarity, we identify the 3D Gaussians relevant to the given text query. To measure volume-aware localization evaluation, we propose a protocol to measure the IoU of 3D Gaussians that expands the traditional metric of point cloud-based approaches by incorporating volumetric information of 3D Gaussians.

\noindentbold{Novel evaluation protocol for 3D localization in 3DGS}
Unlike conventional evaluation protocol for the 3D localization task in point clouds, it is tricky to evaluate 3D localization performance in 3D Gaussians~\cite{3dgs}. This is primarily due to the un-deterministic structure of Gaussian distribution. To address this issue, we compute 3DGS pseudo-labels for evaluating the 3DGS localization in a volume-aware way. The details can be found in the supplementary material.

Given the ground truth, we measure IoU considering the spatial significance of each Gaussian and define a significant score $d_{i}$ for each Gaussian $\theta_i$ with its scale $\mathbf{s}_i = [s_{ix}, s_{iy}, s_{iy}]$ and opacity $\alpha_i$ as $d_{i} = s_{ix}s_{iy}s_{iz} \alpha_i$, where $s_{ix}s_{iy}s_{iz}$ denotes a relative ellipsoid volume of a Gaussian $\theta_i$. With the obtained significant scores $\mathbf{d}=[d_1, d_2, ..., d_{N}]$, we compute weighted IoU of 3D Gaussians to approximate volumes. 
The proposed metric is designed to assign a larger weight to the Gaussians with higher significant scores, when measuring IoU. Figure \ref{fig:significant_score} shows that the impact of each Gaussian on the scene extremely varies depending on their significant scores, which demonstrates the necessity of the proposed IoU metric on 3D Gaussians that regards unequal contributions of each Gaussian. 

\noindentbold{Results}
We report the 3D localization performance on the Scannet dataset in~\Tref{table:3d_loc}. 
The 2D rasterization-based methods~\cite{langsplat, legaussian} struggle to achieve precise activations for 3D localization.
They inherently face challenges when applying for 3D tasks because they need to render 2D images for the scene interaction.
Even with the 3D space search method, OpenGaussian~\cite{open_gaussian}, our model consistently demonstrates superior performance and achieves higher accuracy in localization. 
Figure \ref{fig:6_seg} also shows that LangSplat-m and LEGaussians-m fail to properly localize the objects, and OpenGaussian misses queried objects in the scene.

%% file: section/5_3.tex
\subsection{3D semantic segmentation}
\label{subsec:5_3}

\noindentbold{Settings}
For a given set of open-vocabulary text labels, we perform segmentation by assigning each Gaussian a label having the highest activation among the known label set.

\noindentbold{Results}
The numerical comparison is presented in~\Tref{table:3d_seg}. Although not explicitly designed for semantic segmentation, our model achieves notable performance in this task as a result of accurately updating each Gaussian with language features. Consistent with previous observations, rasterization-based 3DGS models exhibit lower segmentation performance. While OpenGaussian performs position-based clustering, our model demonstrates comparable performance, surpassing the baseline as the Top-$k$ value increases. Our model also achieves better segmentation results, with a visual comparison of the segmented scene shown in~\Fref{fig:6-2-seg}.

%% file: section/5_4.tex
\subsection{Ablation study}
\label{subsec:5_4}
We conduct an ablation study using the ScanNet dataset on different hyper-parameters of \nickname to measure the contribution of each component.

\noindentbold{Product Quantization}
PQ introduces a trade-off between memory usage, computational efficiency, and accuracy. 
To better understand the balance between computational cost and localization quality, we conduct an ablation study by varying the number of sub-vectors. We evaluate performance at sub-vector sizes of 64, 128, and 256. Notably, these settings correspond to bit-size reductions of 1/32, 1/16, and 1/8 of the original CLIP feature, respectively. We measure the query distance computation time for one million data points, averaging results over 100 iterations for efficiency measure. Our findings reveal a favorable trade-off between quantization performance and accuracy (see~\cref{fig:ablation}-(b)) in the Pareto front with our PQ configurations. This achieves a balance that maximizes memory and computational efficiency while minimizing any loss in accuracy.

\noindentbold{Top-$k$ Gaussians}
We examine the influence of the number of Gaussians assigned per ray. This parameter affects both memory requirements and computation, serving as a critical factor in overall performance. The ratio of pruned Gaussians and the mIoU results from different $k$ are presented in~\cref{fig:ablation-a}. We observe that increasing the aggregating number of Gaussians per ray improves localization performance; however, it results in higher memory consumption and the number of occupied Gaussians, indicating a clear trade-off.

%% file: figures/8-ablation/two_graphs.tex
\begin{figure}[t!]
    \centering
    \begin{subfigure}[t]{0.48\linewidth}
        \centering
        \includegraphics[width=1.0\linewidth]{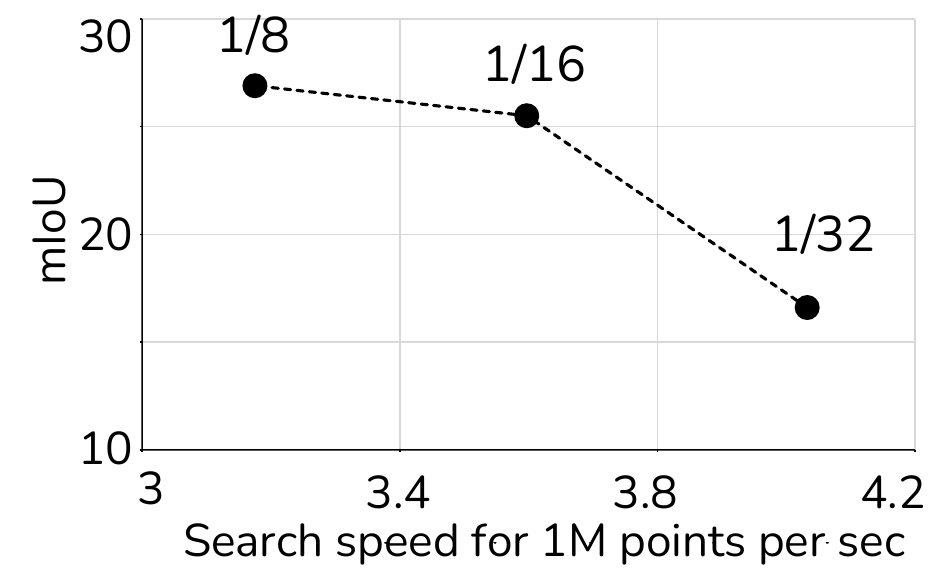}
        \label{fig:subvec}
        \vspace{-4mm}
        \caption{Ablation on PQ parameters.}
        \label{fig:ablation-a}
    \end{subfigure}%
    ~ 
    \begin{subfigure}[t]{0.48\linewidth}
        \centering
        \includegraphics[width=1.0\linewidth]{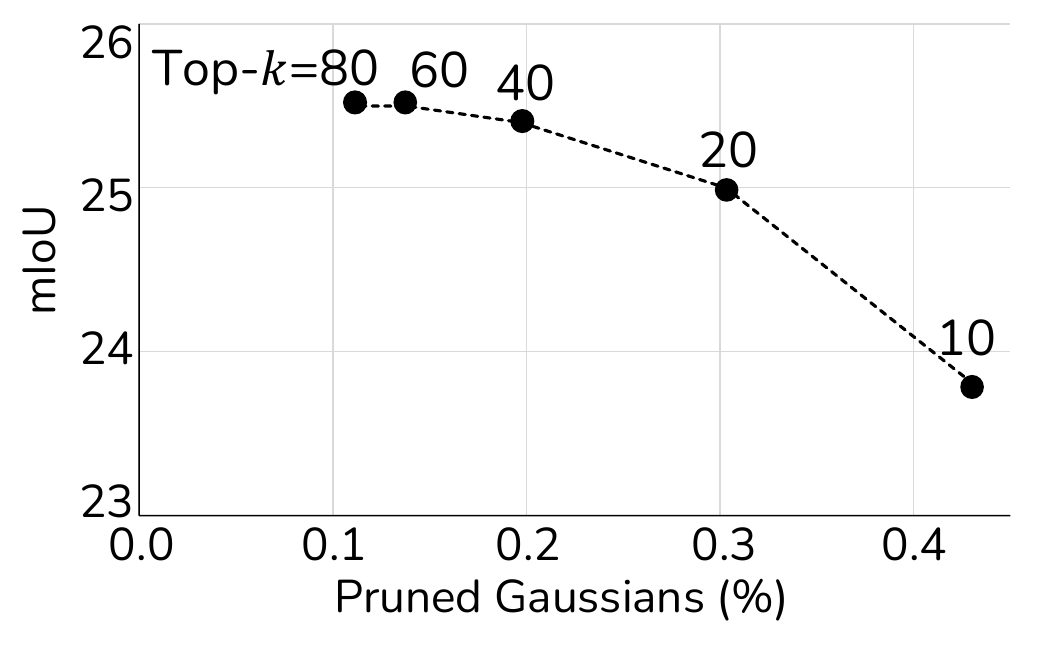}
        \label{fig:topk}
        \vspace{-4mm}
        \caption{Ablation on Top-$k$.}
        \label{fig:ablation-b}
    \end{subfigure}
    \vspace{-2mm}
    \caption{Ablation study on (a) PQ and (b) Top-$k$ Gaussians.}
    \label{fig:ablation}
\end{figure}

%% file: section/6_conclusion.tex
\section{Discussion and Conclusion}
\label{subsec:6_conclusion}
We present \nickname, which is a novel approach for open-vocabulary 3D scene understanding by directly registering language embeddings to 3D Gaussians, eliminating the need for an intermediate rendering process. 
Compared to the previous 2D rendering-based methods~\cite{langsplat,legaussian}, which have limited search domain and capacity, our method directly searches 3D space while preserving the fidelity of language embeddings.
This operation is further accelerated by the integration of Product Quantization (PQ) 

Experimental results validate \nickname’s superior performance across various 3D scene understanding tasks, including open-vocabulary 3D object selection, 3D object localization, and 3D semantic segmentation.
These findings highlight \nickname's ability to transform 3D scene understanding by achieving a balance between highly representative quality and computational efficiency. 
This breakthrough paves the way for advanced applications in robotics, autonomous navigation, and augmented reality.

%% file: supp/0_head.tex
\section*{Supplementary Material}

In this supplementary material, we provide additional details omitted from the manuscript. \Sref{sec:implementation} covers implementation and evaluated 3D tasks. \Sref{sec:exp_setup} outlines the experimental setup, and \Sref{metric} explains our Gaussian-friendly evaluation protocol. \Sref{pq} presents search-time experiments, while \Sref{additional results} includes qualitative results, annotation analyses, and city-scale dataset evaluations. \Sref{discussion} addresses limitations and future directions. We also provide a supplementary video that highlights city-scale experiments.

%% file: supp_figures/S1_pq.tex
\begin{figure*}[t]
    \centering
    \begin{minipage}{0.495\linewidth}
        \includegraphics[width=\linewidth]{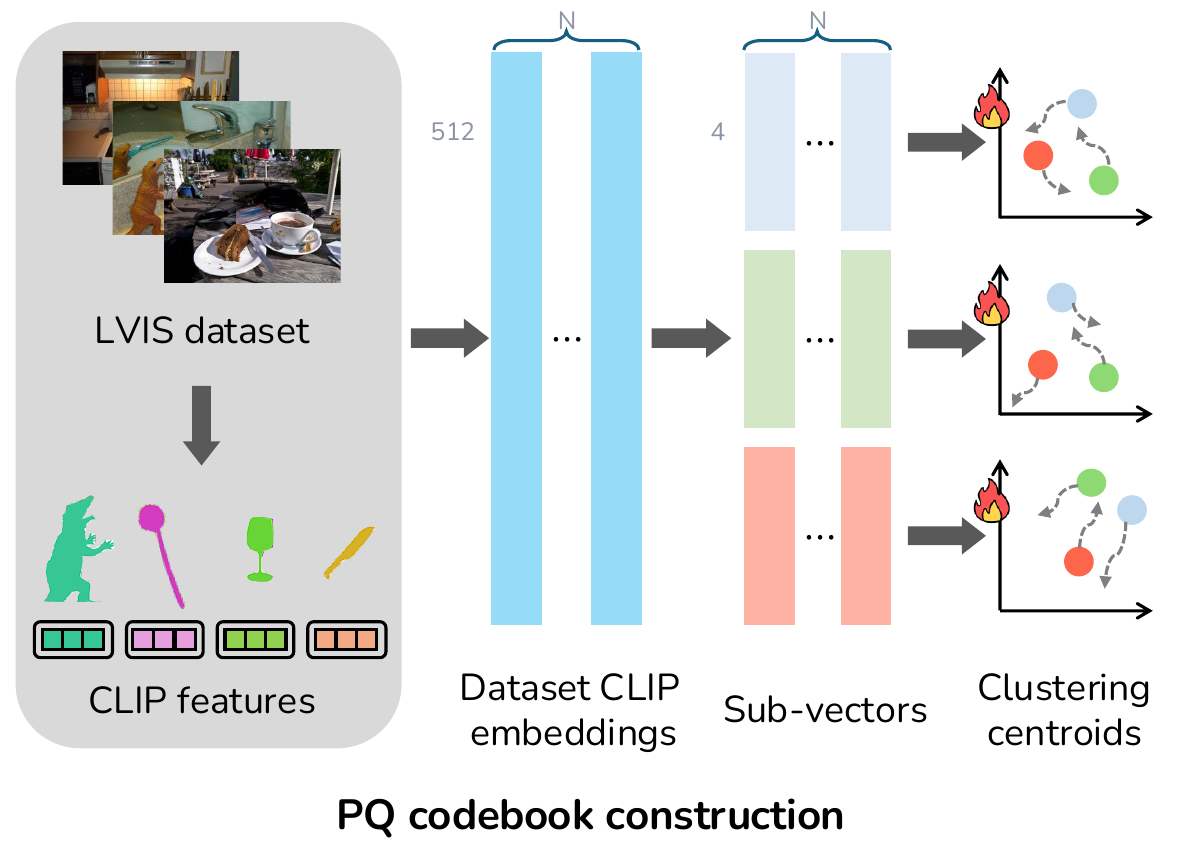}
    \end{minipage}%
    \begin{minipage}{0.01\linewidth}
        \centering
        \begin{tikzpicture}
            \draw [line width=0.5mm, gray] (0, -3) -- (0, 3);
        \end{tikzpicture}
    \end{minipage}%
    \begin{minipage}{0.495\linewidth}
        \includegraphics[width=\linewidth]{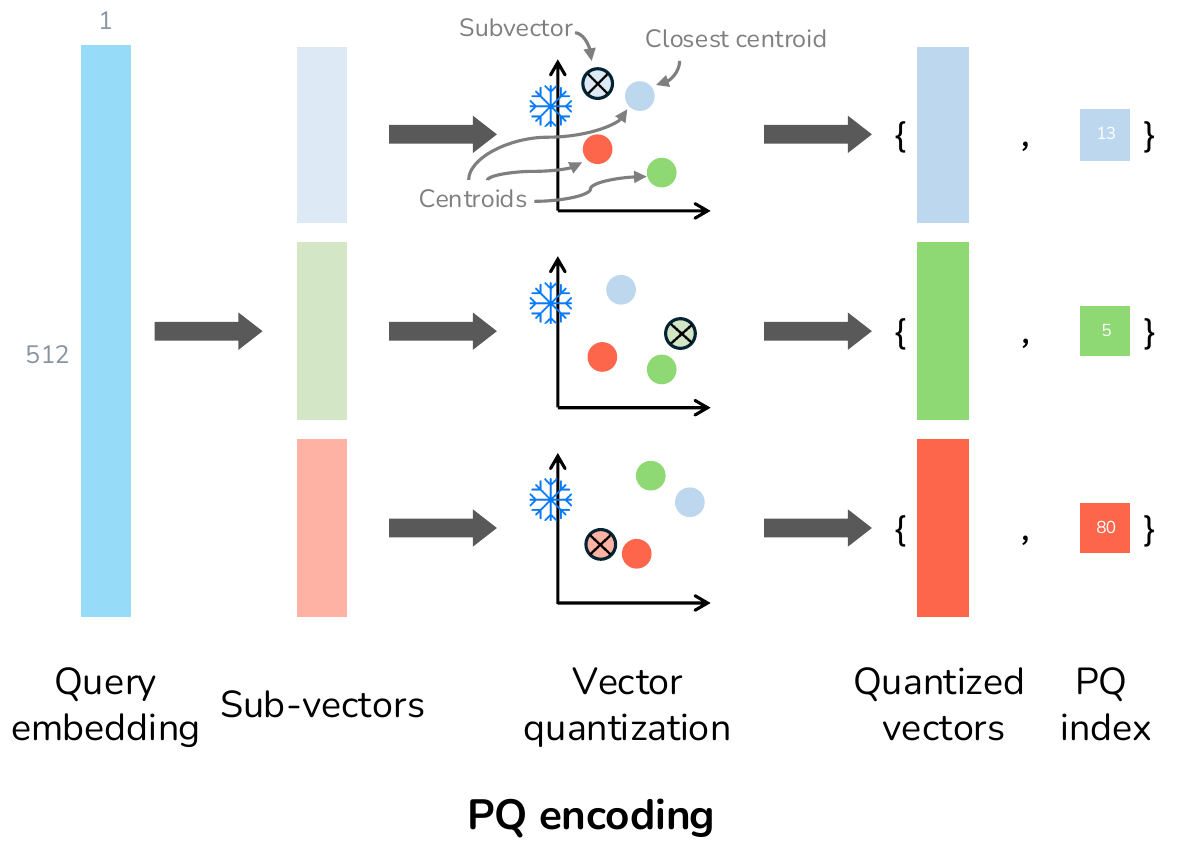}
    \end{minipage}
    \vspace{-3mm}
    \caption{We illustrate the process of construction and encoding process of Product Quantization we used in \nickname. (left) We first construct by update subvector centroids using CLIP features extracted from large-scale object images. (right) After constructing PQ codebook, centroids for each sub-vectors are kept frozen. For query feature, we divide into sub-vectors and are encoded into centroid indices by finding nearest neighbor.}
    \label{fig:product quantization}
\end{figure*}

%% file: supp/1.implementation.tex
\section{Implementation Details}
\label{sec:implementation}
Overall, our method consists of (1) a pre-processing stage that constructs the codebooks in Product Quantization, and pre-training of 3DGS, (2) a training stage that aggregates multiview CLIP embeddings into unified Gaussian-registered embeddings, and (3) Inference stage that directly referring to language-embedded 3D Gaussians for the downstream task.

\paragraph{Pre-Training stage} 
In the pre-processing stage, we need to extract per-patch CLIP embeddings to build PQ codebooks. 
It consists of a patch extraction step, and a CLIP embedding extraction step. 
To obtain patches, we utilize the LVIS dataset, a large-scale dataset having ground truth image segmentations. From the segmentation mask given in the LVIS dataset, we identify object regions and crop them into individual image patches. Each cropped patch is then processed and encoded using the OpenCLIP ViT-B/16 model. Based on these predictions, we continue to build PQ codebooks. We utilize FAISS~\cite{douze2024faiss} open-source library for our Product Quantization implementation.  
We use 128 sub-vectors per embedding, with each subvector assigned to one of 256 centroids, yielding an 8-bit index per subvector. This process is illustrated in \Fref{fig:product quantization}.

3D Gaussian parameters~$\Theta$~\cite{3dgs} are also optimized during the pre-processing stage. We typically care about this initialization of the 3D Gaussians, which can potentially impact the performance of the 3D scene understanding tasks. So, we follow the original 3D Gaussian Splatting method and utilize the optimized 3D Gaussians as our initial parameters. In other words, the pre-training is conducted using the default hyperparameters from 3DGS~\cite{3dgs} framework, running 30,000 iterations. Also, we consistently apply this paradigm across different methods for fair comparison. Especially, for LeGaussian~\cite{legaussian} that employs mutual training, we disabled 3D Gaussian updates during feature assignment in our experiments. 

\paragraph{Training stage}
Based on the PQ and the initial 3D Gaussian parameters~$\Theta$, we begin the training stage. All competing models and our proposed model are trained and evaluated on a single NVIDIA RTX A6000 GPU to ensure fair performance comparison. The training stage consists of three main steps: extracting pixel-wise CLIP embeddings from training images, the feature aggregation stage, and lastly feature registration stage.

Given multi-view images, we extract dense CLIP features assigned to each pixel. 
To obtain per-pixel CLIP features, we adopt the feature extraction scheme by LangSplat~\cite{langsplat}, which utilizes SAM~\cite{sam}. To collect per-patch embeddings, we followed the OpenGaussian framework and used a single-level mask, while LangSplat utilized multi-level masks. 

Once these CLIP features are extracted for all images, we proceed to the feature registration step.
In the feature registration step, we iteratively measure the contribution (weights) of pre-trained Gaussians for each ray assigned to the pixels in the training images, and update the 3D Gaussian embeddings. 
These weights are determined according to the volume rendering equation, which defines their influence during the color rasterization process (see Sec. 3.2 of the manuscript). After the registration process, we normalize the embeddings by dividing embeddings with L2 norms.

Lastly, we register aggregated features to 3D Gaussians. For memory efficiency, we quantize the aggregated Gaussians using the pre-trained PQ codebooks to encode features to indices, a set of 128-channel 8-bit integer indices (\Fref{fig:product quantization}). While registration, Gaussians that were never selected in the top-k process are pruned to reduce noise and memory consumption. At the end of the training, we retain a set of assigned Gaussians with 128 8-bit integer indices.

\paragraph{Inference stage}
Finally, in the inference stage, the PQ-assigned Gaussians from the previous steps are used. By recalling the PQ index list assigned to each 3D Gaussian, cosine similarity is computed between the embeddings of a given text query, extracted using the same CLIP encoder, and each 3D Gaussian. Detailed steps are provided in Sec. 3.3 of the manuscript. As the subvector norms do not sum to 1, normalization by the sum of the subvector L2 norms is applied. We can apply this by using the \texttt{search} function in the Faiss library. The resulting similarity scores are then used to perform various 3D tasks, as evaluated in the study. The following sections explain how the computed activation values are applied in each task.

%% file: supp_figures/S2_metric_qualitative.tex
\begin{figure*}[!ht]
    \centering
        \includegraphics[width=\linewidth]{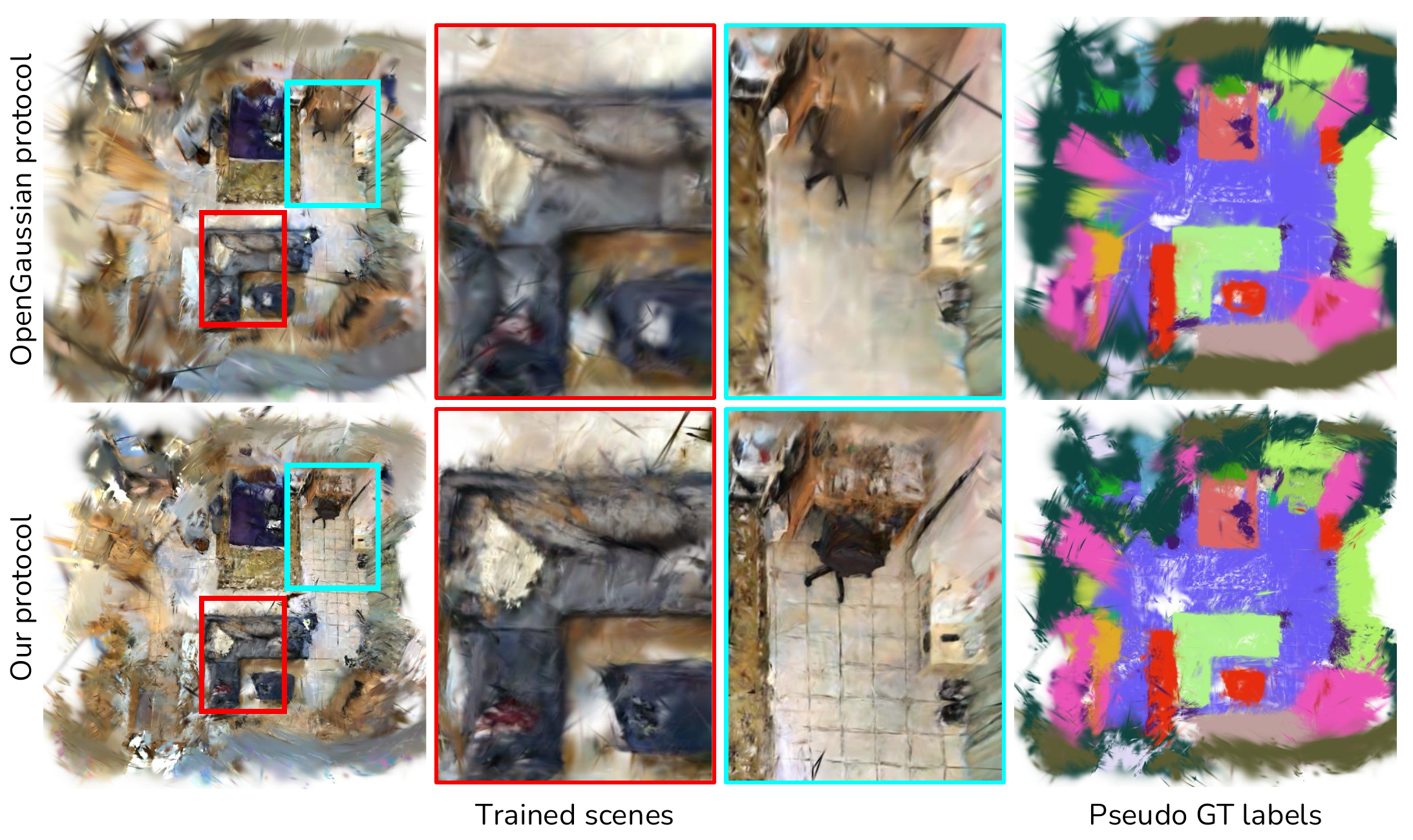}
    \vspace{-3mm}
    \caption{We compare the quality of the scenes and pseudo ground truth labels obtained from different evaluation protocols. (top) Trained scenes following OpenGaussian evaluation protocol, which fix the positions and the number of the initial points during training. (bottom) Trained scenes following our evaluation protocol, which dose not require any constraint during training.}
    \label{fig:metric_qualitative}
\end{figure*}

%% file: supp/2.experiment_setup.tex
\section{Experiment Setup}
\label{sec:exp_setup}

We conduct experiments on three different tasks: 3D object selection task, open-vocabulary 3D object localization task, and open-vocabulary 3D semantic segmentation task. These tasks are closely related to the 3D search as described in Fig. 1 of the manuscript as well as the 3D scene understanding tasks~\cite{open_gaussian}.

\paragraph{3D object selection}
To evaluate the model's 3D awareness capability, we evaluate a 3D object selection task. We first extract text features from an open-vocabulary text query using the CLIP text encoder~\cite{clip}. Next, we compare these text features to the 3D Gaussian embeddings by computing the cosine similarity score. By thresholding the similarity, we identify the 3D Gaussians that are relevant to the given text query. The threshold value for each method is determined through a grid search to identify the optimal performance.

We use the LeRF-OVS dataset~\cite{lerf} with annotations by LangSplat~\cite{langsplat}. As the LeRF-OVS dataset lacks 3D ground truth, we follow the 2D segmentation-based evaluation method proposed by OpenGaussian~\cite{open_gaussian}. This approach evaluates 3D understanding by measuring multi-view 2D segmentation accuracy between the rendered occupancy mask from the selected 3D Gaussians and the GT object masks. Ground truth segmentation masks are manually annotated corresponding to text queries as described in~\cite{langsplat}. We evaluate the IoU and localization accuracy for the metric.

\paragraph{Open-vocabulary 3D object localization}
Given an open-vocabulary text query, we use the CLIP text encoder~\cite{clip} to extract a text feature of the given text query. Then, we compute the cosine similarity score between the query text feature and the Gaussian-registered embeddings. Finally, we select highly relevant 3D Gaussians by thresholding the obtained cosine similarities. We set the threshold of each method individually by searching the thresholds that show the best mIoU on the scenes used for evaluation.

\paragraph{Open-vocabulary 3D semantic segmentation}
We further evaluate our method using the open-vocabulary 3D semantic segmentation task. For a given set of open-vocabulary text queries representing categories, we use the CLIP text encoder to extract a language embedding for each query. We then compute the cosine similarity scores between the 3D Gaussian embeddings and the language features from the given text queries. Using the obtained cosine similarity scores, we assign each 3D Gaussian to the category with the highest the cosine similarity score.

%% file: supp/3.1.prev_eval_protocol.tex
\section{Evaluation Protocols}
\label{metric}

\paragraph{Limitations of existing evaluation protocols}
Compared to the previous works, such as LERF~\cite{lerf}, LEGaussian~\cite{legaussian}, and LangSplat~\cite{langsplat}, our method challenges to leverage the 3D Gaussian representation into the 3D scene understanding tasks. Similar to ours, OpenGaussian~\cite{open_gaussian} is a concurrent work that aims at the open-vocabulary 3D semantic segmentation task as well. However, unlike OpenGaussian, we introduce a new evaluation criterion specialized for the 3D Gaussians, instead of using point cloud-specific evaluations.

OpenGaussian~\cite{open_gaussian} computes evaluation metrics directly from 3D Gaussians, using ScanNet~\cite{dai2017scannet} ground truth point clouds with semantic labels. It aligns Gaussian centers~${ \bmu }$ with dataset points~${ [x,y,z] }$ and keeps both~${ \bmu }$ and the number of Gaussians~$N$ fixed during parameter optimization. This differs from vanilla 3D Gaussian Splatting~\cite{3dgs}. As shown in~\cref{fig:metric_qualitative}, their approach introduces significant quality issues, influenced by the evaluation metric. However, the reason behind this optimization trick is related to evaluation.

The evaluation by OpenGaussian involves predicting labels for each Gaussian and measuring their alignment with the ground truth point cloud using Intersection over Union (IoU). To compute IoU, the overlap (intersection) and total extent (union) of the points are calculated between the 3D Gaussians' center locations~$\{ \bmu \}$ and the ground truth point clouds at fixed positions. As we discussed, since OpenGaussian does not update the locations of the 3D Gaussians, which is identical to the locations of the 3D ground truth points, they simply count the overlap and union without considering the volumetric properties of the 3D Gaussians.

We claim that such an evaluation protocol has two dominant issues. 
First, by pre-defining the number of Gaussians as well as the center locations of the 3D Gaussians, the optimized 3D Gaussians produce degraded rendering quality as shown in~\cref{fig:metric_qualitative}, which is not a practical solution. 
Second, the aforementioned IoU is calculated only with the number of 3D Gaussians, which does not consider the significance of each Gaussian having different shapes and densities.

%% file: supp/3.2.our_eval_protocol.tex
\paragraph{Our Gaussian-friendly evaluation protocol}
To address these limitations, we propose a novel evaluation protocol to compute IoU from 3D Gaussians. Our evaluation protocol follows the original 3D Gaussian Splattings' optimization scheme~\cite{3dgs} by updating the location of the 3D Gaussians as well as the number of 3D Gaussians. After we obtain the optimized Gaussians~$\Theta$, these parameters are used to train language-embedded Gaussians. Then, the following question is how we assign the ground truth semantic labels for each Gaussian from the existing per-point semantic annotations provided by the ScanNet dataset~\cite{dai2017scannet}.

Starting from the given Q numbers of point cloud~$\mathcal{P} = \{ \mathbf{p}_k \}_{k=1}^{Q}$
and a set of semantic labels~$\mathcal{S} = \{ \mathbf{s} \}$,
we compose a paired set of points and their labels as $\{ \mathbf{p}_k, \mathbf{s}^{\mathbf{p}_k} \}_{k=1}^{Q}$,
which is provided by the official datasets. We measure the Mahalanobis distances between the language-embedded 3D Gaussian parameters $\Phi = \{ \theta_i, \mathbf{\tilde f}_i \}_{i=1}^N = \{ \bmu_i, S_i, R_i, \alpha_i, \mathbf{c}_i, \mathbf{\tilde f}_i \}_{i=1}^N$
(Sec. 2 of the manuscript) and ground truth point clouds. Note that the Mahalanobis distances is already used in the 3DGS~\cite{3dgs} when computing effective alpha values, as stated in the Eq. 1 of the manuscript. We maintain to use this equation to calculate the Mahalanobis distance~$\mathbf{d}^\text{mahal}(\cdot)$ between volumetric 3D Gaussian~$\theta$ and 3D point~$\mathbf{p}$ as:
\begin{equation}
    \mathbf{d}^\text{mahal}(\mathbf{p}, \theta) = (\mathbf{p}-\mu)^\top\mathbf{\Sigma}^{-1}(\mathbf{p}-\mu).
    \label{eq:mahalanobis_distance}
\end{equation}
Using the Mahalanobis distance, we determine the semantic label of each Gaussian as below:
\begin{equation}
    \mathbf{s}^{\theta_i} = \mathop{\arg\max}_{\mathbf{s} \in \mathcal{S}} \big( \sum_{\mathbf{p}_k \in \mathcal{P}} \mathbbm{1}{\{ \mathbf{s}^{\mathbf{p}_k} = \mathbf{s} \}} \cdot \mathbf{d}^\text{mahal}(\mathbf{p}_k, \theta_i) \big),
     \label{eq:pseudo_labeling_Gaussians}
\end{equation}
where $\mathbf{s}^{\theta_i}$ is the semantic label of the $i$-th 3D Gaussian~$\theta_i$, $\mathbbm{1}{\{ \mathbf{s}^{\mathbf{p}_k} = \mathbf{s} \}}$ is an indicator function returning $1$ only when $k$-th point label is identical to a semantic label $\mathbf{s} \in \mathcal{S}$. In shorts, this equation determines the semantic label of each 3D Gaussian from the specific semantic label~$\mathbf{s}$ that has the highest sum of the Mahalanobis distances from the ground truth point to each 3D Gaussian.

The proposed assignment process enables generally applicable evaluation of 3D Gaussians without any constraints. \cref{fig:metric_qualitative} shows the quality degradation of the trained scene following the OpenGaussian evaluation protocol, which fixes the position and the number of initial points during training. On the other hand, our generalizable evaluation protocol does not impose any constraints during the training of Gaussians, and it also enables high-quality scene reconstruction, effectively capturing detailed areas.

With the obtained $N$ number of pseudo GT 3D Gaussians, we measure IoU  by considering the volumetric significance of each Gaussian. We define the significant score $d_{i}$ for each Gaussian $\theta_i$ with its scale $\mathbf{s}_i = [s_{ix}, s_{iy}, s_{iy}]^\top$ and opacity $\alpha_i$ as $d_{i} = s_{ix}s_{iy}s_{iz} \alpha_i$ where $s_{ix}s_{iy}s_{iz}$ denotes a relative ellipsoid volume of a Gaussian $\theta_i$. With the obtained significant scores $\mathbf{d}=[d_1, d_2, ..., d_{N}]^\top$, we calculate IoU of $i$-th 3D Gaussians for the label as: 
\begin{equation}
    \begin{gathered}
        \text{Intersection}_i = \mathbf{d}\cdot(\mathbf{l}^{\text{pred}}_i \odot \mathbf{l}_i^{\text{gt}}),\\
        \text{Union}_i = \mathbf{d}\cdot (\mathbf{l}_i^{\text{pred}}+\mathbf{l}_{i}^{\text{gt}}-(\mathbf{l}_i^{\text{pred}} \odot \mathbf{l}_i^{\text{gt}})),\\
        \text{IoU}_i = {\text{Intersection}_i}/{\text{Union}_i},
    \end{gathered}
\end{equation}
where $\mathbf{l}_i^{\text{pred}} \in \mathbb{R}^N$ and $\mathbf{l}_i^\text{gt}\in \mathbb{R}^N$ are binary vectors indicating whether the predicted/GT label of each Gaussian is the $n$-th label,~$\mathbf{s}^{\theta}$ in~\cref{eq:pseudo_labeling_Gaussians}. The proposed metric is designed to assign a larger weight to the Gaussians with higher significant scores when measuring IoU, and the significant score endows our metric with volume-awareness.

\paragraph{Volume awareness of the proposed metric}
To validate that the proposed metric can effectively approximate the volumetric IoU of the 3D scene, we compare our metric with another volume-aware IoU measurement based on voxel representation. Before measuring IoU with voxels, we train 3D Gaussians and generate labeled pseudo-GT 3D Gaussians with \Eref{eq:pseudo_labeling_Gaussians}. Then we first sample voxels in the scene, and allocate a GT label to each voxel with the labeled 3D Gaussians. We obtain the most likely label of each voxel by defining the label score. The label score $l_{jn}^\text{voxel}$ is computed with the opacity $\alpha_i$ and the density $\mathcal{N}(\mathbf{v}_j | \bmu_i, \mathbf{\Sigma}_i)$ of each Gaussian at the position of a voxel $\mathbf{v}_j$ as:
\begin{equation}
    l_{jn}^\text{voxel} = \sum_{\theta_i \in \Theta}\alpha_i \cdot \mathbbm{1}\{\mathbf{s}^{\theta_i} = \mathbf{s}\} \cdot \mathcal{N}(\mathbf{v}_j | \bmu_i, \mathbf{\Sigma}_i),
    \label{eq:pseudo_labeling_voxels}
\end{equation}
where $\mathbbm{1}\{\mathbf{s}^{\theta_i} = \mathbf{s}\}$ is an indicator function determining whether a Gaussian $\theta_i$ is assigned to the $n$-th label and $\text{det}(\mathbf{\Sigma}_i)$ is the determinant of $\mathbf{\Sigma}_i$. With the obtained score, we first filter out empty voxels by thresholding with: $p_j = \sum_{n=1}^{L}l^\text{voxel}_{jn}$, where $L$ is the total number of the labels, which can be interpreted as a density of each voxel $\mathbf{v}_j$. Then we assign a label with the highest score, as the GT label of each voxel. We can also generate predicted labels of voxels using the predicted labels of Gaussians in the same manner, and can evaluate IoU by comparing the GT and predicted labels of the voxels one-to-one.

Volume awareness is inherent in this voxel-based IoU evaluation as the voxels explicitly represent the volume of the scene. We show the volume-awareness of our evaluation metric by showing a correlation between our metric and voxel-based metric in \Fref{fig:metric_correlation}. As can be seen, our metric obtains a high correlation with the voxel-based IoU evaluation metric by considering the significant score when calculating IoU. This result shows the necessity of the significant score, which endows our metric with volume awareness.

Although the voxel-based IoU evaluation effectively measures volume-aware IoU of the scene, the computational cost to assign labels is too expensive. Each time new labels of Gaussians are predicted, the process of assigning them to the voxels is required for evaluation. Different from voxel-based IoU evaluation, our IoU evaluation protocol has a low computational cost, since there is no repeated assignment process after we once generate the labeled pseudo-GT Gaussians. In other words, our proposed IoU evaluation protocol is a fast and volume-aware evaluation for measuring the IoU of scenes represented by 3D Gaussians.  

%% file: supp_tables/metric_comparison.tex
\begin{table}[t]
    \centering
    \resizebox{1.0\linewidth}{!}{
                \begin{tabular}{l|cc|cc}
                    \toprule

                                          & \multicolumn{2}{c|}{OpenGaussian evaluation} &  \multicolumn{2}{c}{Our evaluation} \\
                    \multicolumn{1}{c|}{} &  OpenGaussian &Ours  &  OpenGaussian&Ours \\ 
                    
                    \midrule
                    IoU $>$ 0.15          &  52.7 &\textbf{54.3}  & \textbf{57.8} & 52.6  \\
                    IoU $>$ 0.30          &  36.4 &\textbf{39.4}   & 38.0 & \textbf{40.3} \\
                    IoU $>$ 0.45          &  14.7 &\textbf{15.5}   & 18.3 & \textbf{25.6} \\
                    \midrule
                    3D mIoU         &  23.1 &\textbf{25.0}   & 25.2 & \textbf{25.4} \\
                    % \midrule
                    % $>$ 0.15 & IoU $>$ 0.3 & IoU $>$ 0.45 \\
                    \bottomrule

                \end{tabular}
    }
    \caption{We compare different metrics for measuring IoU, proposed by OpenGaussian~\cite{open_gaussian} and our work.
    } 
    \label{table:metric_comparison}
\end{table}

%% file: supp_figures/S3_metric_correlation.tex
\begin{figure}[h!]
    \centering
        \includegraphics[width=\linewidth]{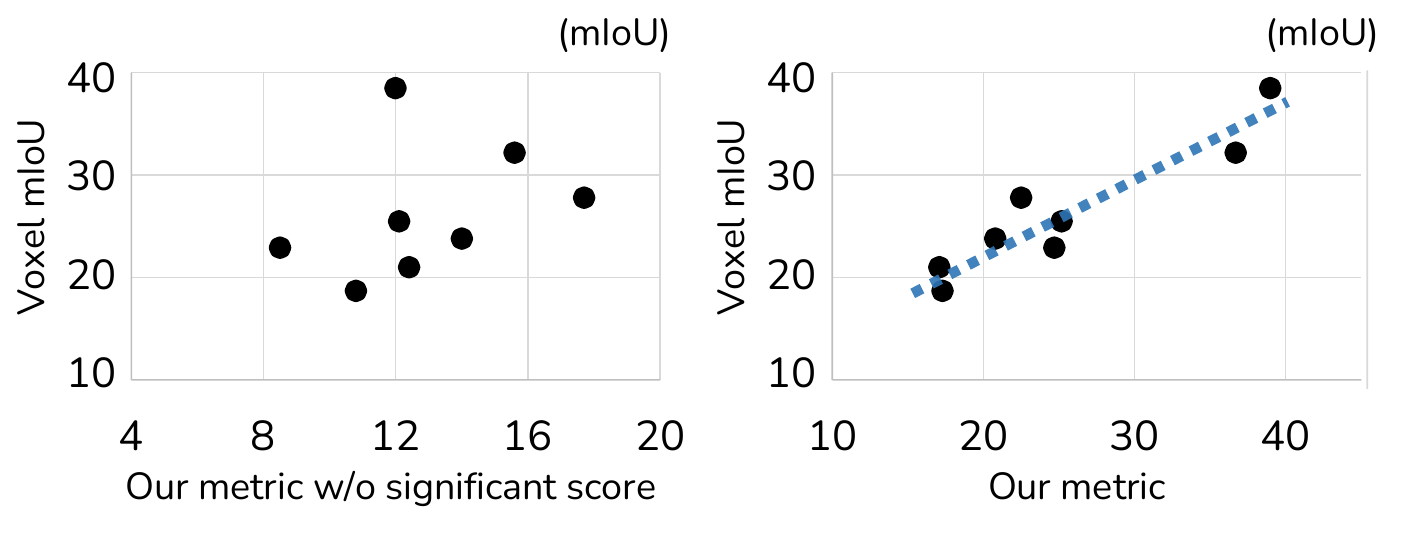}
    \vspace{-3mm}
    \caption{Scatter plot of mIoUs with different mIoU evaluation protocols, measured from eight scenes of the ScanNet~\cite{dai2017scannet} dataset. (left) Low correlation between voxel-based metric and our metric without significant score, \ie, same score $d_i$ for all Gaussians. (right) High correlation between voxel-based metric and our metric.}
    \label{fig:metric_correlation}
\end{figure}

%% file: supp/4.pq.tex
\section{Search-time Experiments}
\label{pq}

In addition to its memory efficiency, Product Quantization significantly enhances search speed. Product quantization can approximate distances between vectors using quantized sub-vectors. By precomputing and storing distances between subvector centroids in a Look-Up Table (LUT), distance calculations between query and database vectors during the search phase are reduced to simple indexing operations. The precomputation shifts the complexity of vector distance calculations from $O(ND)$ for a $D$ dimensional vector to $O(N)$ per subvector.

Use of LUT can be described as follows. For trained PQ centroids $c_{lj}$, for $l = 1, 2, \dots L$, and $j= 1, ..., 2^k$, where $L$ is number of sub-vectors, and $k$ refers the number of bits used for indexing each centroids. LUT is stored as follows:
\begin{equation}
    \textrm{LUT}_{l}[i, j] = ||c_{li}-c_{lj}||_2^2, \textrm{where}\quad i, j \in \{1, 2, ... 2^k\}. 
\end{equation}
Then for vectors, $\mathbf{v}_1 = [\mathbf{v}_{11}, \ldots, \mathbf{v}_{1L}]$, $\mathbf{v}_2 = [\mathbf{v}_{21}, \ldots, \mathbf{v}_{2L}]$ mapped to indices $j_1=[j_{11}, j_{12}, ...,j_{1L}]$, and $j_2=[j_{21}, j_{22}, ...,j_{2L}]$, distance is computed as summation of each retrieved LUT values following each PQ indices:
\begin{equation}
    d(v_1, v_2) = \sum_{l=1}^L{\textrm{LUT}_l(j_{1l}, j_{2l})}.
\end{equation}

We can also compute the cosine similarity of the vectors, by computing inner products rather than distances, following normalization by the sum of each norm of each sub-vector. Despite the quantization errors, previous literature~\cite{product_quantization} shows that these errors remain within certain quantization bounds, preserving the correlation between the approximated and actual distances. 

The scalability and speed of the proposed approach make it particularly suitable for handling complex 3D data.
We compared search speed between computing cosine similarity of CLIP features and distance computation in product quantization (see~\Fref{fig:inf_speed}). Under identical hardware conditions, the proposed LUT-based approach demonstrated substantial speed improvements compared to cosine similarity computation between CLIP features: with a subvector size of 128, 64, 32  search performance improved by approximately 2$\times$, 6.6$\times$, 14.1$\times$ respectively.
These improvements underscore the computational advantages of the proposed method.

Considering that rendering-based methods require significantly greater computation compared to 3D data processing, \nickname's approach demonstrates its superior efficiency in search efficiency, and establishes itself as a practical and scalable solution for 3D data search and processing at scale.

\input{supp_figures/S4_inference_speed}

%% file: supp_figures/S4_inference_speed.tex
\begin{figure}[h!]
    \centering
        \includegraphics[width=\linewidth]{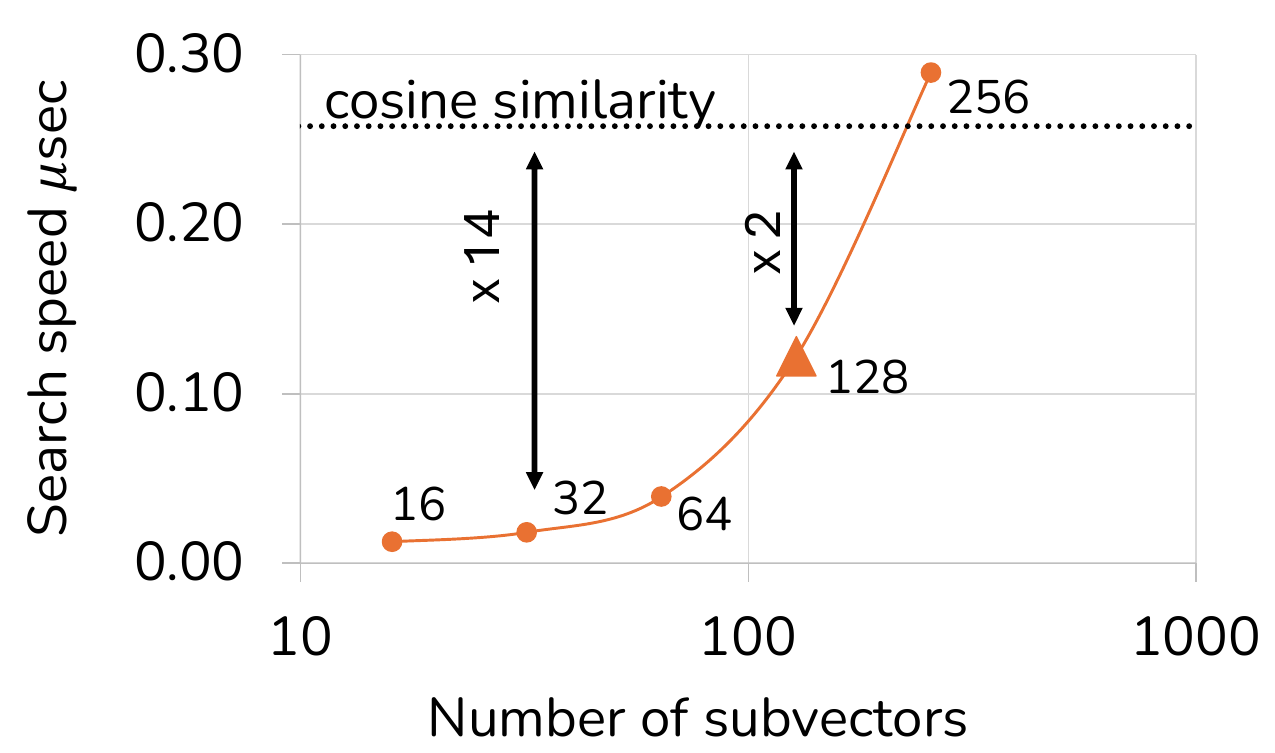}
    \caption{We compare inference speed between product quantization LUT based method and ordinary cosine similarity calculation. We calculate average inference time spent over one million feature points. We report mean values over 100 repeated experiments}
    \vspace{-3mm}
    \label{fig:inf_speed}
\end{figure}

%% file: supp/5_additional_results.tex
\section{Additional Results}
\label{additional results}
In this section, we present additional results that are not shown in the manuscript due to space constraints. 

\subsection{Additional results on presented 3D tasks}
\label{3D tasks}
We first show more experimental results for the 3D object selection task in~\Fref{fig:3d_obj_sel}, and the 3D localization task in ~\Fref{fig:3d_loc} which are not included in the manuscript due to the space limit. Consistent with our earlier observations, the LangSplat model struggles to learn accurate 3D features. While it occasionally follows feature patterns, it frequently produces significant noise, making it unsuitable for real-world applications such as localization, object grabbing, or 3D image editing. Additionally, we observe persistent spatial bias in the OpenGaussian method, as previously noted, see red cup, plate, or wavy noodles, and bed cases in~\Fref{fig:3d_loc}, it fails to select relevant regions in others.
In contrast, our proposed method, which allows direct search and inference in 3D space, consistently identifies favorable localization performance. This demonstrates the robustness and practicality of our approach compared to competing methods.

In the Sec.\ref{metric}, we demonstrated that our metric provides superior volumetric alignment compared to existing approaches. To further validate the superiority of our model, we also evaluated its performance using the metric proposed by OpenGaussian. We confirm that our method outperforms even using other evaluation protocols as shown in~\Tref{table:metric_comparison}

\input{supp_figures/S5_3d_object_selection}
\input{supp_figures/S6_3d_localization}

\subsection{Experiments on the ScanNet200 dataset}
\label{scannet200}
The proposed model and its counterparts are designed to operate effectively in open-vocabulary settings. 
To evaluate performance under more comprehensive open-vocabulary cases, we conducted additional experiments using the ScanNet-200 annotation~\cite{scannet200}, which extends the ScanNet limited-label of 20 to 200 semantic categories, including tail categories such as armchair and windowsill. These rare classes provide a closer approximation to real-world scenarios and enable a robust assessment of the models' generalization capabilities. For consistency, experiments are conducted using the same scenes as previous benchmarks, following ground truth annotations as described in Sec.~\ref{metric}. 

The results, summarized in~\Tref{table:scannet_200}, demonstrate that the proposed model consistently outperforms its counterparts, which highlights superior generalization across diverse object spaces. The results validate the proposed model's ability to excel across both constrained and diverse object spaces, emphasizing its potential for practical application in complex real-world scenarios.

\input{supp_tables/scannet_200}

\subsection{Experiments on the city-scale dataset}
\label{city scale}
\input{supp_figures/large_scene/S7_langsplat_comp}
\input{supp_figures/large_scene/S8_green_red_comp}
\input{supp_figures/large_scene/S9_large_scene}

The proposed method is further evaluated in a large-scale scenario using the Waymo San Francisco Mission Bay dataset~\cite{tancik2022blocknerf}, which features expansive spatial contexts. For each scene, the dataset comprises approximately 12,000 images captured by 12 cameras, providing a challenging and diverse testing environment for 3D localization tasks. We select 3 blocks of the scene for large-scale scene tests.

We conducted comparisons against the LangSplat-m model for the 3D text-query localization task as shown in~\Fref{fig:langsplat_comp}. Our evaluation focused on qualitatively assessing how well each model performs in localizing queries within the 3D space. Our method consistently succeeds in localizing diverse text queries, demonstrating robust and accurate performance across various contexts. In contrast, LangSplat-m struggles to make precise predictions, particularly with its 3D Gaussian representations failing to align with the expected ground-truth values. These findings are consistent with our earlier observations regarding the limitations of LangSplat-m's approach.

As shown in~\Fref{fig:green_red_comp}, we can see that the results reflect not only objects, but also attributes like color to some extent. Additional visualizations of the results can be found in~\Fref{fig:large_scene} and the supplementary video, which provides a more comprehensive view of the qualitative differences between the methods. We strongly encourage readers to refer to these supplementary materials for further insights.

The differences between the methods become even more pronounced when considering search speed in large-scale scenarios. 
For example, the Waymo dataset contains over 2.9M Gaussians, with individual images requiring nearly 1M computations per image for over 100 images. The computational efficiency of the proposed method allows it to handle such large-scale data more effectively, highlighting its scalability and practical applicability in real-world scenarios.

%% file: supp_figures/S5_3d_object_selection.tex
\begin{figure*}[h!]
    \centering
        \includegraphics[width=\linewidth]{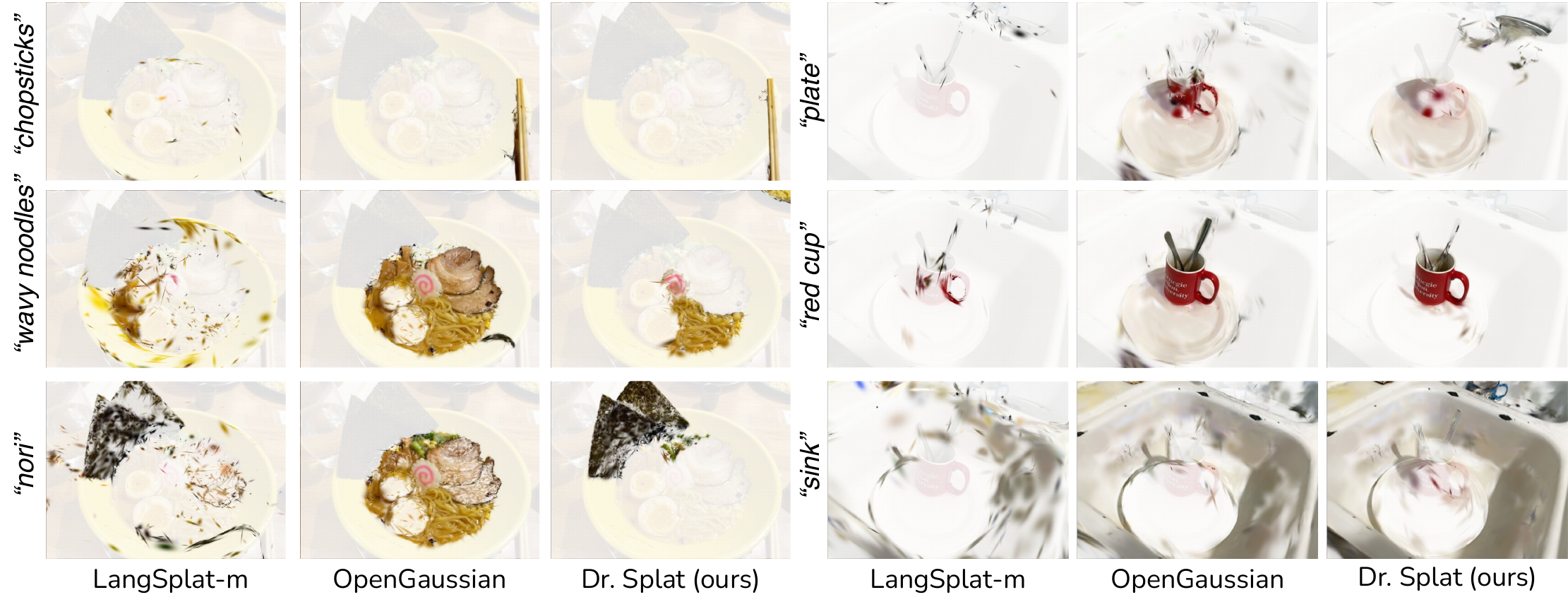}
    \caption{We compare 3D object selection task in LeRF dataset with Langsplat~\cite{langsplat}, and OpenGaussian~\cite{open_gaussian}. We visualize selected Gaussians with high similarity to query text. Langsplat shows noisy, 3D uncorrelated activations, and Opengaussian often show false positive activations, while our method show accurate localization showing superiority on generalizability.}
    \vspace{-3mm}
    \label{fig:3d_obj_sel}
\end{figure*}

%% file: supp_figures/S6_3d_localization.tex
\begin{figure*}[h!]
    \centering
        \includegraphics[width=\linewidth]{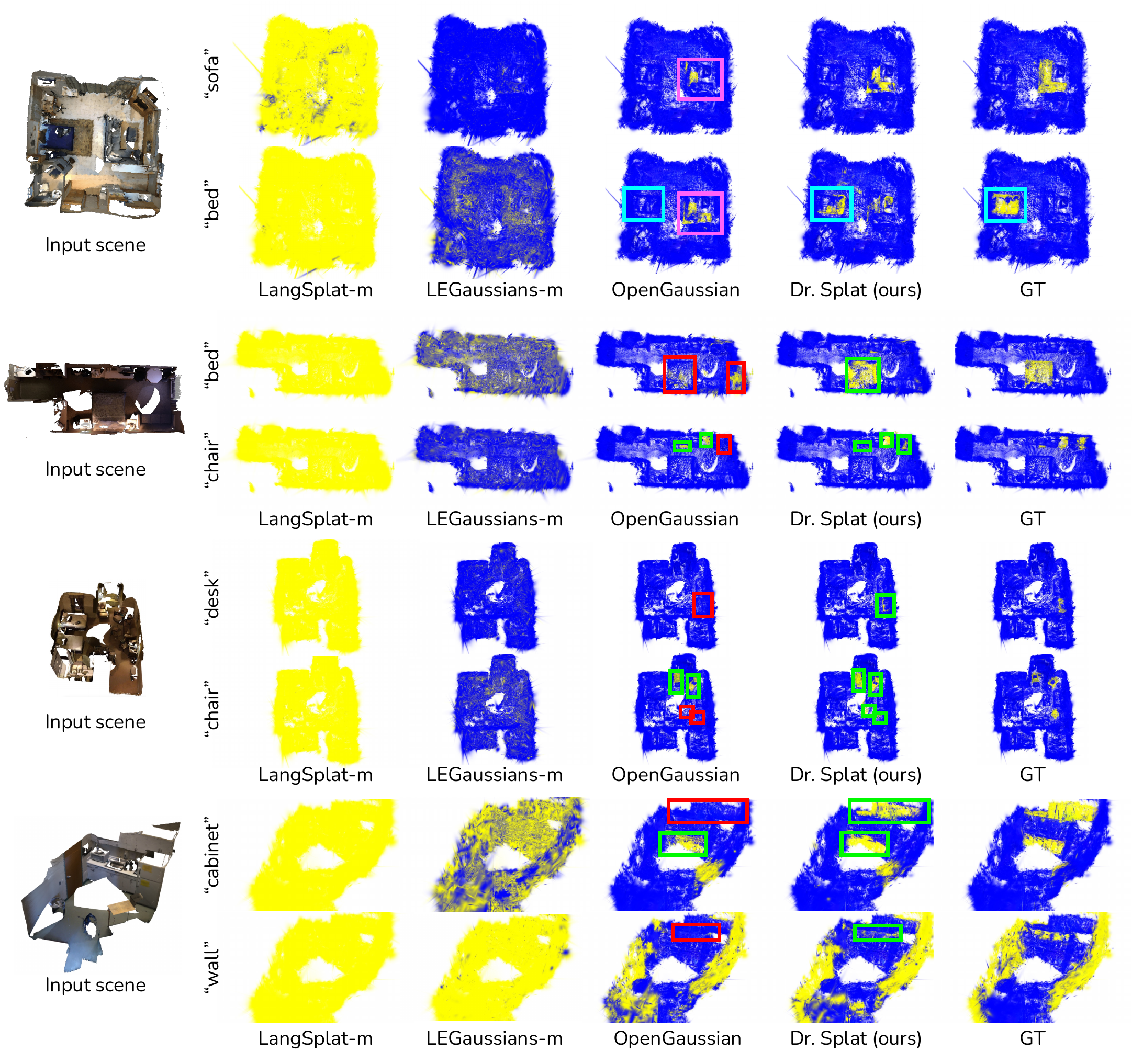}
    \caption{We compare 3D object localization results between competing methods~\cite{langsplat, legaussian, open_gaussian} with \nickname.
    3DGS with similarity above the threshold (0.562) are shown in yellow, while those below the threshold are displayed in blue.
    Greenbox indicates successful localization, while red boxes indicates missing or false positive in 3D localization.}
    \vspace{-3mm}
    \label{fig:3d_loc}
\end{figure*}

%% file: supp_tables/scannet_200.tex
\begin{table}[t]
    \centering
    \resizebox{1.0\linewidth}{!}{
                \begin{tabular}{l|c|ccc}
                    \toprule

                                          & 3D   & \multicolumn{3}{c}{200 classes} \\
                    \multicolumn{1}{c|}{} & mIoU & IoU $>$ 0.15 & IoU $>$ 0.3 & IoU $>$ 0.45 \\
                    \midrule
                    LangSplat-m~\cite{langsplat}              & 3.9  & 7.6 & 3.5 & 0.8 \\
                    LEGaussians-m~\cite{legaussian}          & 4.0  & 7.4 & 3.8 & 1.4 \\
                    OpenGaussian~\cite{open_gaussian}          & 14.7 & 34.2 & 18.9 & 11.0 \\
                    Ours (Top-20)        & 14.6 & \textbf{36.3} & 18.6 & 9.4 \\
                    Ours (Top-40)        & \textbf{ 14.9}  & 36.0 & \textbf{19.3} & \textbf{14.0} \\
                    \bottomrule
                \end{tabular}
    }
    \caption{We compare evaluate our method with previous methods on the ScanNet-200 dataset.
    } 
    \label{table:scannet_200}
\end{table}

%% file: supp_figures/large_scene/S7_langsplat_comp.tex
\begin{figure*}[t!]
    \centering
    \includegraphics[width=\linewidth]{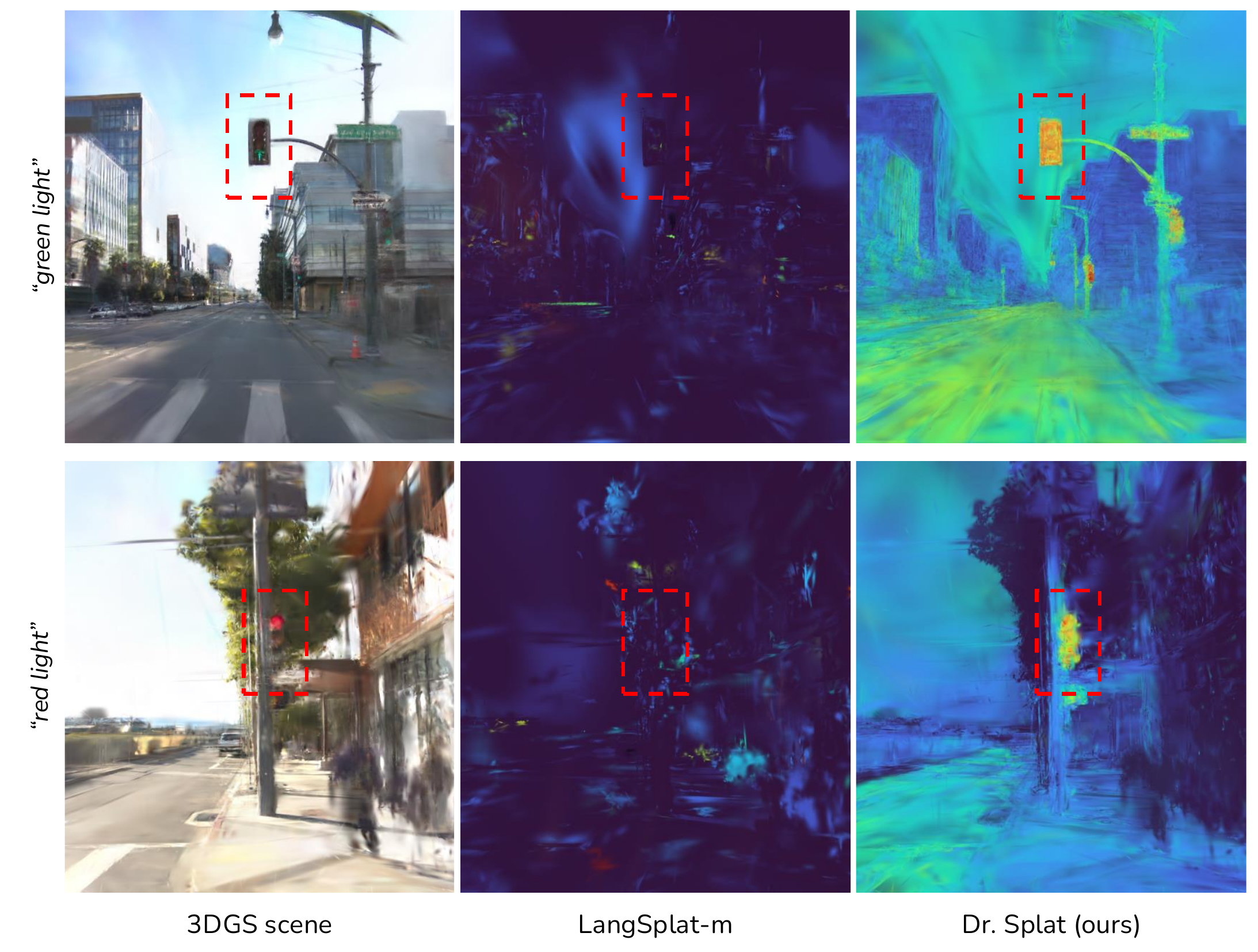}
    \vspace{-6mm}
    \caption{
        We compare 3D localization between rendering based Langsplat-m with registration based \nickname. While LangSplat-m shows randomly distributed activations, fail to localize the target, ours model successfully detect the target in both cases.
    }
    \label{fig:langsplat_comp}
\end{figure*}

%% file: supp_figures/large_scene/S8_green_red_comp.tex
\begin{figure*}[t!]
    \centering
    \includegraphics[width=\linewidth]{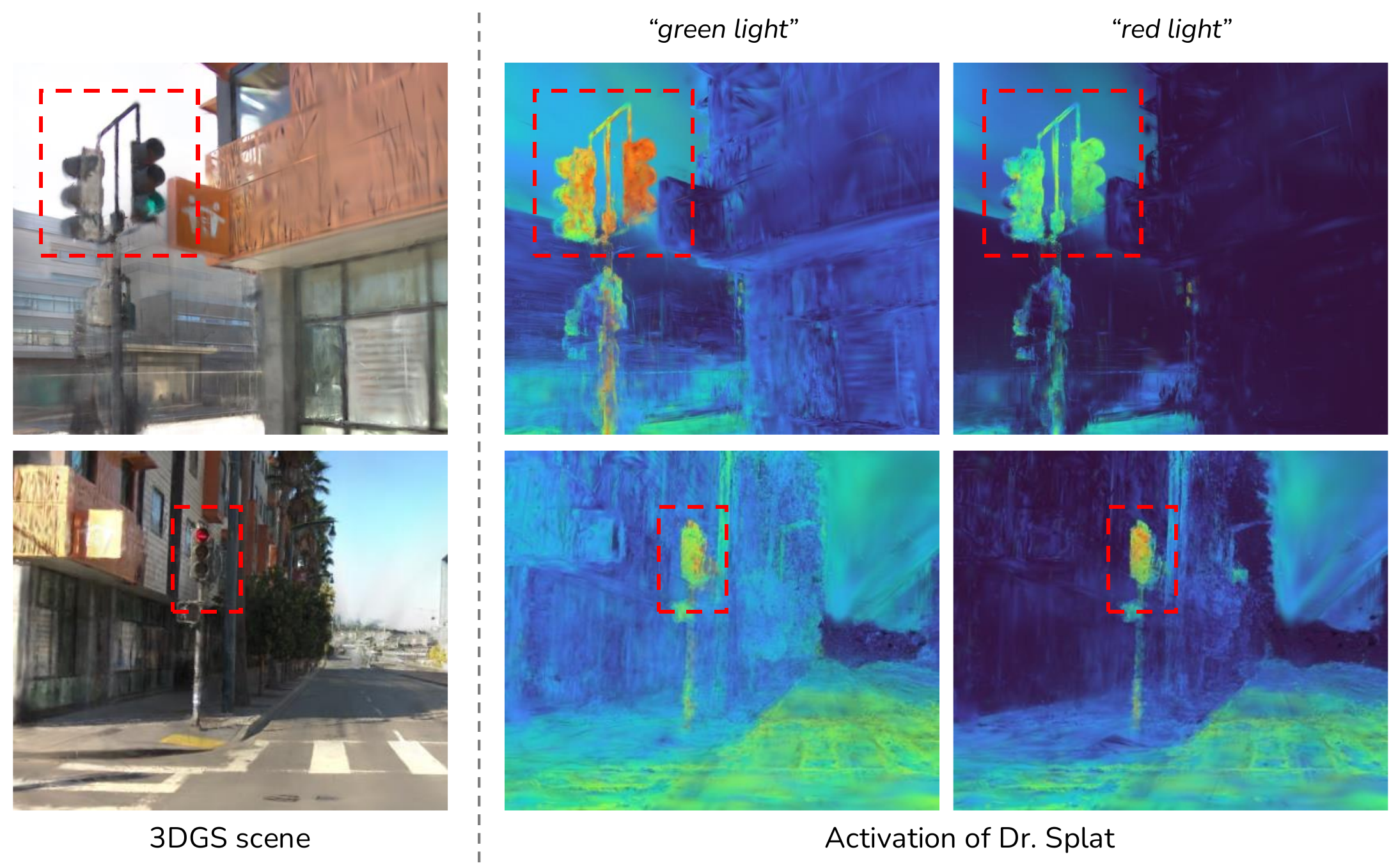}
    \vspace{-6mm}
    \caption{
        Visualization of 3d localization in different attributes (e.g., color) given as query. The result highlights the ability of Dr. Splat (ours) to effectively distinguish attributes such as ``green light'' and ``red light'' in scenes based on text queries, demonstrating the robustness in open-vocabulary understanding.
    }
    \label{fig:green_red_comp}
\end{figure*}

%% file: supp_figures/large_scene/S9_large_scene.tex
\begin{figure*}[t!]
    \centering
    \includegraphics[width=\linewidth]{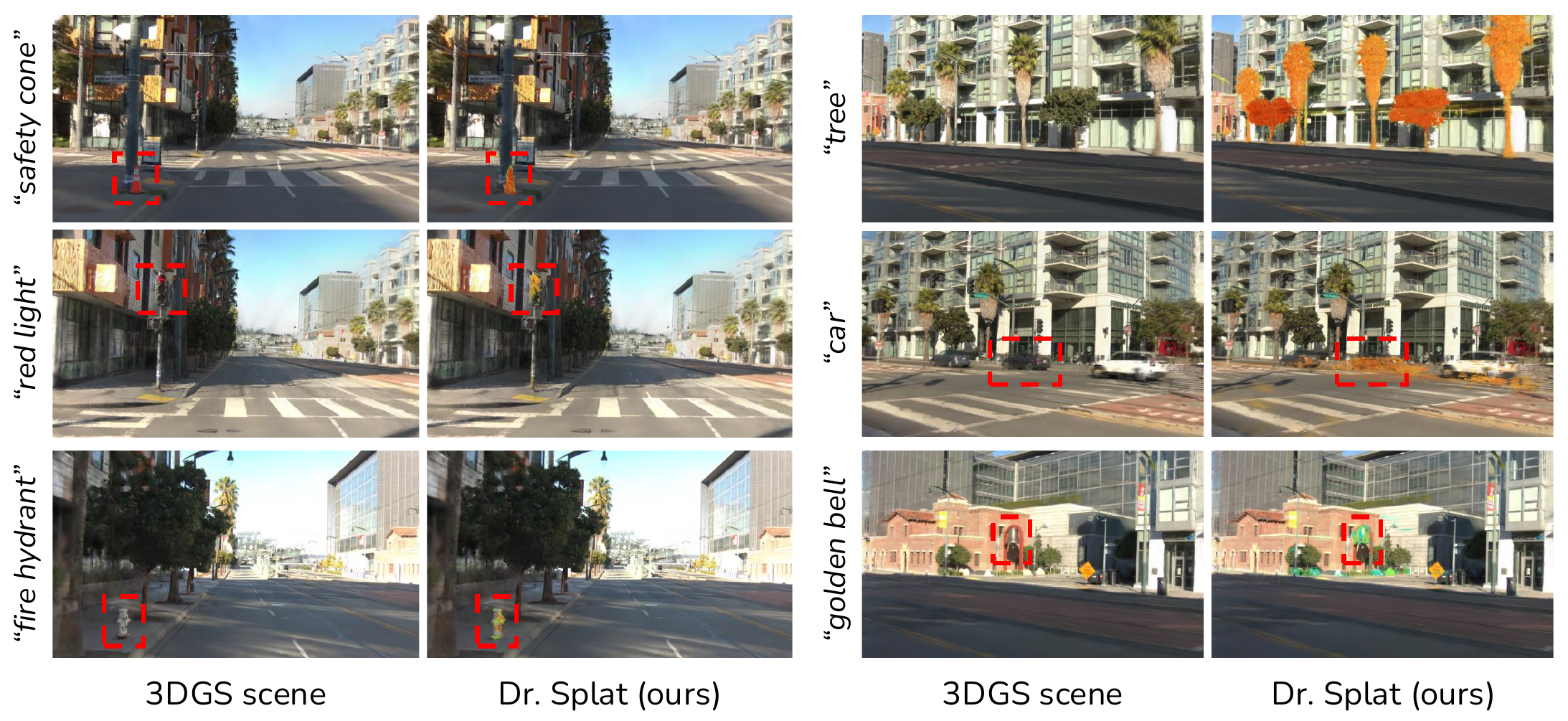}
    \vspace{-6mm}
    \caption{
        Qualitative results of \nickname on 3D localization task in city-scale data showcasing \nickname's generalization performance across diverse text queries includes various target objects and concepts.
    }
    \label{fig:large_scene}
\end{figure*}

%% file: supp/6_limitation.tex
\section{Broader Applications and Limitations}
\label{discussion}
\paragraph{Broader application}
The proposed method offers the potential for broader applications across diverse scenarios. 
Similar to works that explore the application in point cloud~\cite{concept_fusion, open_scene} and MLP-based methods~\cite{lerftogo2023}, our approach, using 3DGS, can be extended to support various input modalities, such as click or image queries, by leveraging a self-referencing mechanism. Additionally, integrating our method with Large Language Models (LLMs) could facilitate dialogue-based interactions, allowing users to dynamically issue commands or explore the environment.
This integration suggests promising avenues for developing 3D interactive systems that go beyond simple search tasks.

Furthermore, applying the method to canonical forms could support dynamic 3D scenes~\cite{deformable3dgs, gagaussian}. This adaptation would extend the applicability of our approach beyond static environments, demonstrating its versatility in handling complex, real-world scenarios.

\paragraph{Limitation}
While our method has demonstrated robust performance across diverse combinations of nouns and adjectives (\eg, ``tea in a glass,'' wavy noodles,'' and red light'' in~\Fref{fig:3d_obj_sel} and~\Fref{fig:large_scene}) as well as unfamiliar nouns (e.g., nori,'' waldo,'' and safety cone''), without additional training, generalization remains an area for improvement. Exploring additional training techniques for Product Quantization (PQ) could further enhance the method’s capabilities. 
Further exploration of Product Quantization (PQ) training, such as using more diverse datasets or finer-grained query representations, could enhance adaptability across varied contexts.

Despite its advantages, some limitations of the proposed method have also been identified, particularly related to CLIP features. Occasionally, related but distinct objects are simultaneously activated for a given query. For instance, the query ``red apple'' might activate non-red apples or unrelated red objects. This stems from CLIP's semantic associations and could be mitigated with post-processing techniques like re-ranking to improve query specificity.

Lastly, similar to previous methods~\cite{langsplat, legaussian, open_gaussian}, ours also requires to set an appropriate threshold. In this study, we utilize a fixed similarity threshold employed a fixed similarity threshold across all scenes, ensuring stable and reproducible results. However, optimizing thresholds for specific scenarios or implementing dynamic adjustments could further refine localization accuracy in diverse environments.